\documentclass{article}

\usepackage{arxiv}

\def\eg{\emph{e.g}\onedot}

\usepackage[utf8x]{inputenc} 
\usepackage[T1]{fontenc}    
\usepackage[pagebackref,breaklinks,colorlinks]{hyperref}       
\usepackage{url}            
\usepackage{booktabs}       
\usepackage{amsfonts}       
\usepackage{amsmath}
\usepackage{amssymb}
\usepackage{nicefrac}       
\usepackage{microtype}      
\usepackage[capitalize]{cleveref}       
\usepackage{lipsum}         
\usepackage{graphicx}
\usepackage{natbib}
\usepackage{doi}

\usepackage[linesnumbered,ruled,vlined]{algorithm2e}\DontPrintSemicolon
\usepackage[toc]{appendix} 
\usepackage{boldline,multirow}
\usepackage{xcolor}
\usepackage{caption}
\usepackage{subcaption}
\usepackage{enumitem}

\SetKwComment{Comment}{// }{}

\newcommand{\assign}{\leftarrow}
\newcommand{\var}{\texttt}
\newcommand{\FuncCall}[2]{\texttt{\bfseries #1(#2)}}
\SetKwProg{Function}{function}{}{}

\definecolor{color_init}{RGB}{82, 62, 218}
\definecolor{color_pers}{RGB}{255, 187, 92}
\definecolor{color_target}{RGB}{179, 230, 150}

\newcommand\T{\rule{0pt}{2.6ex}}       

\crefname{section}{Sec.}{Secs.}
\Crefname{section}{Section}{Sections}
\Crefname{table}{Table}{Tables}
\crefname{table}{Tab.}{Tabs.}

\title{FindView: Precise Target View Localization Task for Look Around Agents}

\date{March 23, 2022}

\author{\href{https://orcid.org/0000-0003-1494-3635}{\includegraphics[scale=0.06]{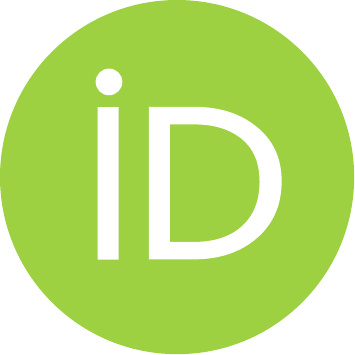}\hspace{1mm}Haruya Ishikawa}\thanks{\href{https://github.com/haruishi43}{github.com/haruishi43}} \qquad Yoshimitsu Aoki \\
	Department of Electrical Engineering \\
	Keio University\\
	Yokohama, Kanagawa 223-0061 \\
	\texttt{haruyaishikawa@keio.jp} \\
}

\renewcommand{\shorttitle}{FindView Task Benchmark for Look Around Agents}

\hypersetup{
pdftitle={FindView: Precise Target View Localization Task for Look Around Agents},
pdfsubject={},
pdfauthor={Haruya Ishikawa, Yoshimitsu Aoki},
pdfkeywords={Reinforcement Learning, Look Around Agents, Computer Vision},
}

\begin{document}
\include{pythonlisting}

\maketitle

\begin{abstract}

With the increase in demands for service robots and automated inspection, agents need to localize in its surrounding environment to achieve more natural communication with humans by shared contexts.
In this work, we propose a novel but straightforward task of precise target view localization for look around agents called the FindView task.
This task imitates the movements of PTZ cameras or user interfaces for $360^\circ$ mediums, where the observer must ``look around" to find a view that exactly matches the target.
To solve this task, we introduce a rule-based agent that heuristically finds the optimal view and a policy learning agent that employs reinforcement learning to learn by interacting with the $360^\circ$ scene.
Through extensive evaluations and benchmarks, we conclude that learned methods have many advantages, in particular precise localization that is robust to corruption and can be easily deployed in novel scenes.

\end{abstract}


\section{Introduction}
\label{sec:intro}

\begin{figure}[t]
\centering
\includegraphics[width=0.4\linewidth]{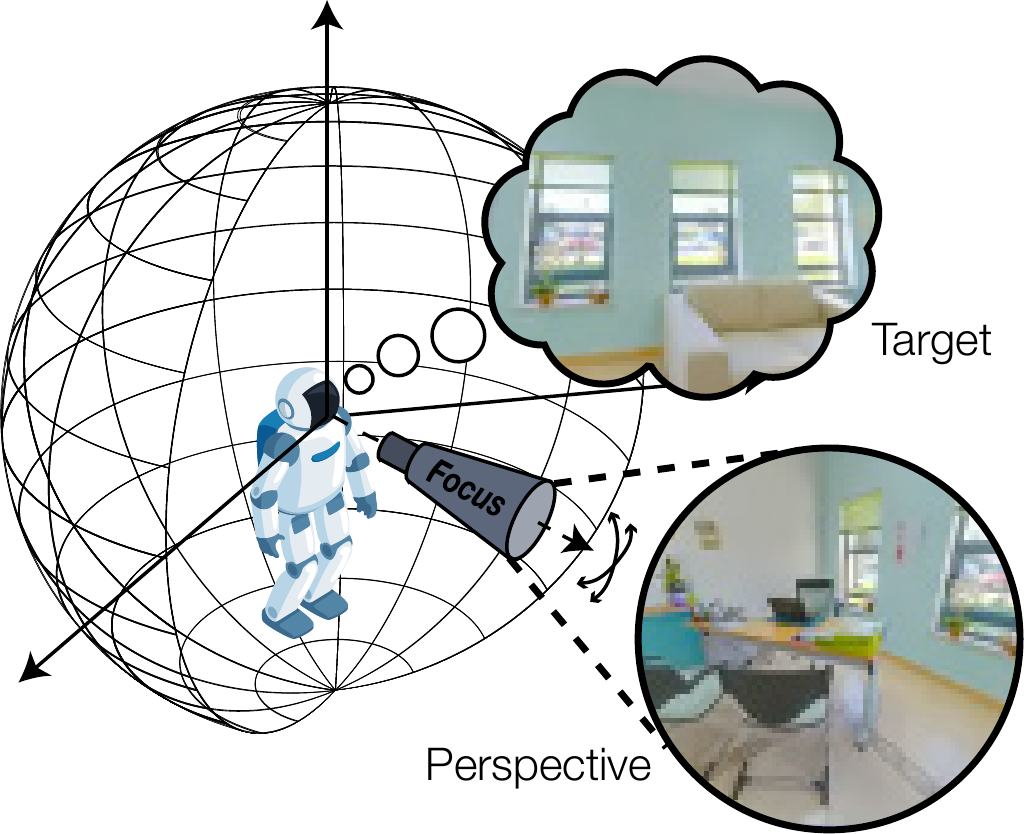}
\caption{In the FindView task, the agent is tasked to ``look around" and find a specific target view in the scene. The agent needs to find similarities in its' perspective views to understand the scene and localize the target view.}
\vspace{-1.2em}  
\label{fig:overview}
\end{figure}

\noindent\textbf{Embodied Agents.}
Over the recent years, there have been increased attention to visual navigation and embodied agents \cite{zhu2017target,savva2019habitat,kolve2019ai2thor,ActiveNeuralLocalization,du2021vtnet,hahn2021no,shah2021ving}. 
The field of research aims to create agents that use visual sensors for solving complex tasks or aid humans by learning to perceive, communicate, and act in their environment.
Humans in the loop make the goal very difficult since the dynamics of the environment are changeable, and human interactions can lead to unexpected events.
Towards better collaboration between agents and humans, agents must be able to perform localization of any point in space that reflects the characteristics of human's perception of 3D space \cite{cirik2020refer360}.
Since the visual sensors for the agents are commonly RGB sensors employed with partial Field-of-View (FoV), we would need to train these agents to perceive how humans see from these views.
Communication with these agents will almost always necessitate the agents to navigate to view a common referential FoV in the scene so that the human can instruct the agents with the shared contexts.
Challenge arises since the point of interest could be any point in the scene, and many points in the scene will not correspond to easily named objects.
So far, many embodied agents being researched use either partial FoVs or directly use panoramic images that are hard for human observers to understand.
We believe that embodied agents should be able to look around and localize in various views that human observers might be looking at.
We approach this problem by introducing a new task, namely the FindView task, to evaluate and benchmark the agents (\cref{fig:overview}).

\noindent\textbf{Pan-Tilt-Zoom Cameras.}
Nowadays, Pan-Tilt-Zoom (PTZ) cameras are used for a variety of applications such as surveillance, video conferencing, live production, lecture capture, and distance learning \cite{lalonde2007system,zhang2020deepptz}.
PTZ cameras are installed on walls or ceilings, and the observers control the viewing direction and FoV of the camera to their liking.
The freedom to view anywhere in the $360^\circ$ scene makes these cameras more appealing in many industries, especially for remote surveillance of public areas or secluded factories.
The utilization of PTZ cameras gained significance as people’s free movement became largely hindered by the prevalence of the COVID-19 pandemic.
A recent example is Boston Dynamics Spot, which has an onboard PTZ camera for remote surveillance and inspection.
Following its success, we can observe an urgent need to develop user-friendly viewing interface to foster smoother communication with these types of robots and other autonomous agents.
In our work, we mainly target agents, like PTZ cameras, that look around in $360^\circ$ to find a view of interest.

\noindent\textbf{User-Friendly $\bf{360^\circ}$ Content Viewing.}
Despite the increasing number of head-mounted displays (HMD), many $360^\circ$ images and videos are still being viewed by users on existing 2D displays, which are commonly available on websites such as Flickr, Youtube, and Facebook.
A large proportion of the $360^\circ$ videos are filmed virtual tours, with some even live streaming.
The $360^\circ$ images are commonly used in showcasing real-estate listings, so users can look around the house without being present there.
Although much previous research has made interactions with these mediums easier (\eg picture-in-picture) \cite{lin2017outside,li2021route}, new research has taken one step further and shown how the automatic creation of normal FoV (NFoV) views can minimize user interactions \cite{su2016pano2vid,su2017making,lai2017semantic,hu2017deep,lee2018memory,kang2019interactive,wang2020transitioning360}.
For example, generating hyperlapses \cite{lai2017semantic} for $360^\circ$ videos is an innovative means of summarizing long footage of $360^\circ$ video and showcasing some of the monumental moments automatically.
By adopting automatic methods, we can gain wider access to contents at the compromise of human control, thus able to provide better contents accessibility for the visually impaired \cite{iwamura2020visphoto}.
In our work, we aim to create an agent that can find similar views by looking around.
We believe that this could enable a wide array of applications such as enabling users to find a view in a $360^\circ$ image or video that closely resembles some NFoV view. 


\noindent\textbf{Overview of the Paper.}
As expressed through the contexts of embodied agents, PTZ cameras, and $360^\circ$ contents, we believe that there is a need for models that can localize precisely and efficiently by looking around.
We propose a simple, but novel task of precise target view localization for look around agents, namely the FindView task.
We have introduced two agents for solving this task: rule-based and policy learning agents.
Through extensive evaluations and benchmarks, we conclude that there are many advantages to learned methods.

\section{Related Works}
\label{sec:related_works}


\noindent\textbf{Visual Localization.}
Visual localization, coined in \cite{larsson2019fine}, is the problem of estimating the camera pose of a given image relative to a visual representation of a known scene.
Robust estimation of camera pose is necessary for various applications such as AR and robotics, as well as employing 3D reconstruction methods such as SfM and Visual SLAM for global localization.
For local localization, feature matching methods are heavily studied \cite{lowe2004distinctive,rublee2011orb,sattler2018benchmarking}.
Robust matching methods have been developed with recent advances in deep learning \cite{sarlin2020superglue,sun2021loftr}.
In our work, we consider local feature matching as a baseline for active view localization.
Our works closely relate to active pose estimation using visual sensors as in \cite{ActiveNeuralLocalization,parisotto2018global}, but this task tries to solve given a global map, which we do not consider for the FindView task.


\noindent\textbf{Visual Navigation.}
For visual navigation, agents (\eg robots) use visual sensors such as RGB cameras to maneuver around in a scene to accomplish given task(s).
With the recent advent of CG simulators, it is easier to train navigational agents in a reinforcement learning manner, and accelerators like GPU enable training in batches \cite{zhu2017target,savva2019habitat,kolve2019ai2thor}.
Following these works, we create a highly efficient and parallelizable simulator for the FindView task to enable training for millions of steps.

\noindent\textbf{Image-Goal Navigation.}
One of the prevalent tasks in visual navigation is the task of image-goal navigation \cite{zhu2017target,savva2019habitat,hahn2021no,shah2021ving}.
In this task, the agent is given a target image and must move around in the scene to find the position that obtains the target image.
Similar to our task, the agent must learn to stop when the target image is acquired.
Standard image-goal navigation tasks consider no movement in pitch directions which heavily limits the view.
Recently, there have been works that consider an incremental movement of $15^\circ$ in the pitch direction \cite{du2021vtnet}.
Since the application for this task is not restricted to robotics, we create the task of precise target localization with fine-grained rotational increments.


\noindent\textbf{Look Around Agents.}
Visual exploration using a ``look-around" agent in $360^\circ$ scenes has been studied in \cite{jayaraman2018learning,ramakrishnan2018sidekick,ramakrishnan2019emergence}.
In \cite{jayaraman2018learning}, they introduce the task of active observation completion where an agent selects camera motions to efficiently reconstruct viewgrid images using a limited set of NFoV glimpses.
A viewgrid is a pseudo $360^\circ$ image where each grid is a portion of the $360^\circ$ image.
They showed that policies learned on this task transferred well to the active categorization task.
In \cite{ramakrishnan2018sidekick}, sidekick policy learning is introduced to improve the performance.
It uses the full $360^\circ$ images to aid the training by mitigating the effects of partial state observability.
In \cite{ramakrishnan2019emergence}, it is shown that the policy learned via sidekick policy learning generalizes well to a range of active perception tasks, including pose estimation.
Our work is similar to the above works for (1) both use perspective images from the $360^\circ$ environment, and (2) both adopt a reinforcement learning approach for policy learning.
However, our work differs in (1) the FindView task uses directions as action space instead of coordinates, (2) the FindView task has around $43,560$ possible views per episode (compared to $24$ views for pose estimation task in \cite{ramakrishnan2019emergence}), and (3) in our work, agents have to determine when to stop for evaluation instead of an exploration budget.



\begin{figure}[ht]
\centering
\includegraphics[width=0.95\linewidth]{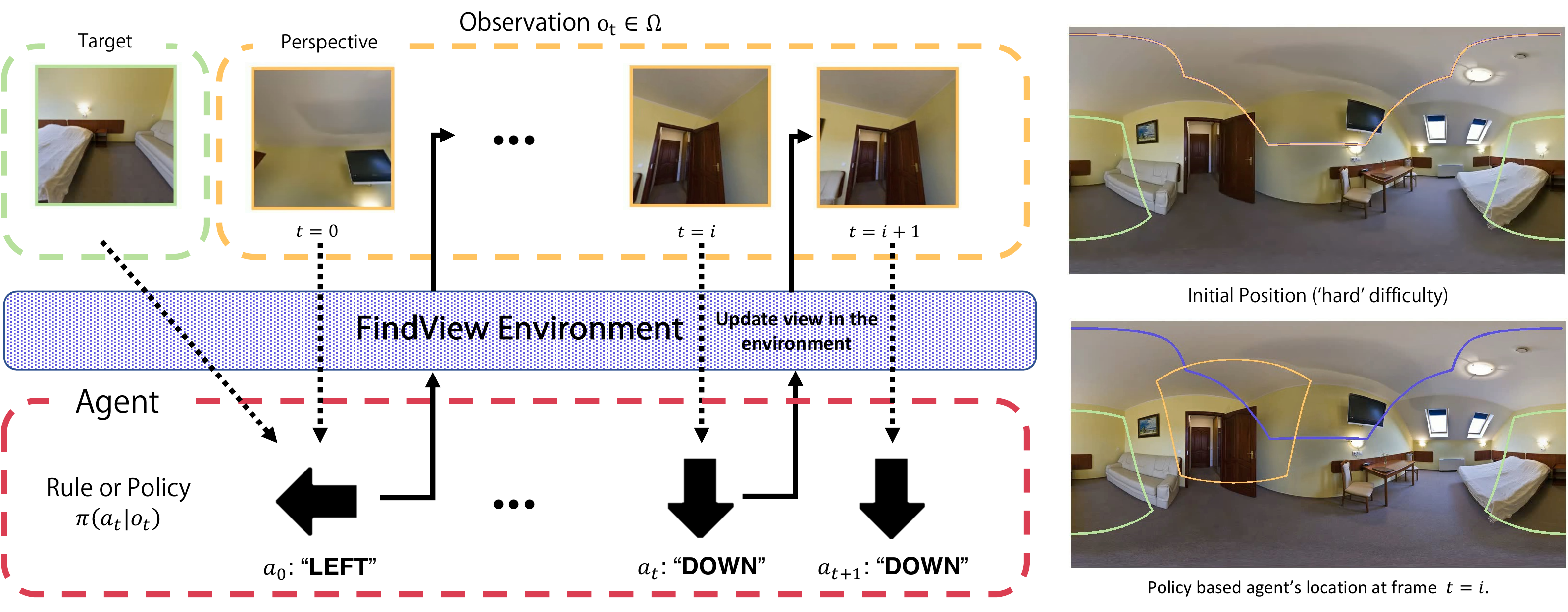}
\caption{The diagram represents the FindView task and the environment. On the right, we have added visualization where the initial, target, and perspective views are projected on the equirectangular image (\textcolor{color_init}{blue}, \textcolor{color_target}{green}, and \textcolor{color_pers}{orange} respectively). First, a target image an initial perspective view is given to the agent. Based on the observation $o_0$, the agent performs action $a_0$. The environment is updated and the next observation $o_1$ is given to the agent. This process repeats until the agent calls \texttt{stop}, or is terminated by the environment.}
\vspace{-1.2em}  
\label{fig:findview_task}
\end{figure}

\section{FindView Task}
\label{sec:findview_task}


Suppose you want to find out how an image you took of your neighborhood in the past appears on Google Street View now.
After searching the address on the map and opening Street View, you can then ``look around" to find a view closely resembling the image.
This is the essence of the FindView task we introduce in this paper.
In this task, the agent is put in a novel scene and has full control over the viewing direction in pitch and yaw movements.
The agent is then given a target image that the agent needs to locate in the scene by ``looking around".
We primarily focus on the pitch and yaw movements because, as stated in \cref{sec:intro}, many applications exist for agents that can automatically look around.

\subsection{FindView Environment}
\label{sec:findview_environment}


The purpose of the FindView task is to benchmark the agent's capabilities of looking around and finding the target view.
A simulator is created to retrieve perspective images given the viewing directions of the agent.
We denote the simulator's FoV as $f=90^\circ$ and perspective image size of $(h=256, w=256)$ for the rest of the paper.
Note that the observation image and target image are of the same size.
See \cref{apdx:simulator} for more information about the simulator.

\cref{fig:findview_task} represents an overview of the FindView task.
The agent is first given a target image and an initial observation.
Observation and target images are RGB images with the same dimensions.
For simplicity, we bundle target image and observation image at step $t$ into $o_t$. 
The agent does not know the rotation $\mathbf{R}_t=(\theta_t, \psi_t)$ ($\theta_t$ and $\psi_t$ denotes pitch and yaw respectively) for neither the initial observation $\mathbf{R}_0$ or target $\mathbf{R}_{\text{target}}$.
The agent is also limited to the observation and cannot see anywhere else until the agent performs some action.

The action is made up of directional movements (\texttt{up}, \texttt{down}, \texttt{left}, \texttt{right}) and \texttt{stop}.
For each step $t$, the agent performs a single action and receives a new observation based on the agents' current viewing direction $\mathbf{R}_t$.
The agent can only move $\delta = 1^\circ$ for the desired direction.
For example, if at step $t$, the agent was viewing $\mathbf{R}_t=(\theta_t, \psi_t)$ and decided to look \texttt{up}, the new rotation would be $\mathbf{R}_{t+1} = (\theta_t + \delta, \psi_t)$.
If the agent decided to then look \texttt{right}, the new rotation would be $\mathbf{R}_{t+2} = (\theta_t + \delta, \psi_t + \delta)$
The value of $\delta$ is hidden from the agent.
The range of the possible pitch movements are $[-\theta_{\text{bound}}, \theta_{\text{bound}}]$, where as the possible movements for yaw are $(-180^\circ, 180^\circ]$.
We have top and bottom boundaries for the pitch direction $\theta_{\text{bound}} = 60^\circ$  which prevents the agent from wrapping around the top and bottom that can cause unwanted roll orientation.
The agent can wrap around in the yaw direction.

Agent must find and call \texttt{stop} within a limited number of steps denoted by the terminal step size of $T=5,000$.
The agent is forced to stop otherwise.

\subsection{Evaluation Metrics}
\label{sec:findview_evaluations}

After the agent has stopped, we evaluate the agent on the metrics explained below:

\noindent\textbf{Localization Error ($\bf{\varepsilon}$).}
To evaluate how precise the agent is at localizing, we evaluate the error in terms of distances.
We take the absolute $\ell^1$ angular distance of the rotation at the final step $t$, which is given by:
\begin{equation} \label{eq:error}
\varepsilon_{i} = \| \mathbf{R}^{i}_{\text{target}} - \mathbf{R}^{i}_t \|_{1} = |\theta^{i}_{\text{target}} - \theta^{i}_{t}| + |\psi^{i}_{\text{target}} - \psi^{i}_{t}|.
\end{equation}
For the final metric, we calculate $\varepsilon = \sum^{N}_{i}\varepsilon_{i}$ where $N$ is the total number of episodes.

\noindent\textbf{Frequency of Calling \texttt{stop} ($\bf{\omega_{\text{stop}}}$).}
This measures the ability to call \texttt{stop} of an agent which is calculated by taking the total number of times the agent has called \texttt{stop} and normalizing it with the total number of episodes evaluated.

\noindent\textbf{Frequency of Perfect Localization ($\bf{\omega_{\text{perf}}}$).}
Given that an agent successfully called \texttt{stop}, we determine the success when the localization error is $0$ for the episode.
We calculate the sum of the successes $N_{\text{perf}}$ and normalize it with the total number of episodes that have been stopped by the agent $N_{\text{stop}}$.

\noindent\textbf{Success Weighted by Path Length ($\bf{\eta}$).}
Since this is a navigation task, we consider the widely adopted Success weighted by (normalized inverse) Path Length (SPL) \cite{anderson2018evaluation}.
The SPL is given by,
\begin{equation} \label{eq:spl}
\eta = \frac{1}{N}\sum^N_i S_i \frac{L^{\text{oracle}}_i}{\max(L_i, L^{\text{oracle}}_i)},
\end{equation}
where in the $i$-th episode, $L_i$ is length of the path the agent took and $L^{\text{oracle}}_{i}$ is the shortest path.
$S_i$ is a binary indicator of success which is given by,
\begin{equation} \label{eq:s_i}
    S_i = 
\begin{cases}
    1,& \varepsilon_i = 0 \\
    0,& \varepsilon_i > 0
\end{cases}.
\end{equation}

\subsection{FindView Dataset}
\label{sec:findview_dataset}

\begin{table}[t]
\small
\caption{Summary of the FindView Dataset.}
\centering
\begin{tabular}{c | c | c}
  \hline\hline
  \makebox[0mm]{Dataset} & \makebox[10mm]{Scene} & \makebox[40mm]{\# of Images (Train/Val/Test)}\\
  \clineB{0-2}{2.0}
  \multirow{2}{*}{SUN360}
                      & Indoor         & 14,358 (11,487/1,432/1,439) \\
                      & Outdoor        & 52,938 (42,352, 5,295, 5,291)\\
  \clineB{0-2}{2.5}
  360-Indoor          & Indoor         & 3,335 (2,668/334/333) \\
  \hline
  \hline
\end{tabular}
\label{table:dataset}
\vspace{-1.2em}  
\end{table}

In the FindView simulator described in \cref{sec:findview_environment}, we use an equirectangular image as the source image to simulate the look around movement.
We used the popular SUN360 dataset to create our indoor and outdoor FindView dataset \cite{xiao2012recognizing}.
We have also used the recent 360-Indoor dataset \cite{chou2020360}.
We have split each dataset into `train', `validation', and `test' where `train' and `validation' were used for training and validating the agents and `test' was used to benchmark the agents.
The statistics for each dataset are shown in \cref{table:dataset}.

\subsection{Difficulties}
\label{sec:findview_difficulties}

\begin{figure}[ht]
\centering
\begin{subfigure}[b]{0.3\textwidth}
  \centering
  \includegraphics[width=\textwidth]{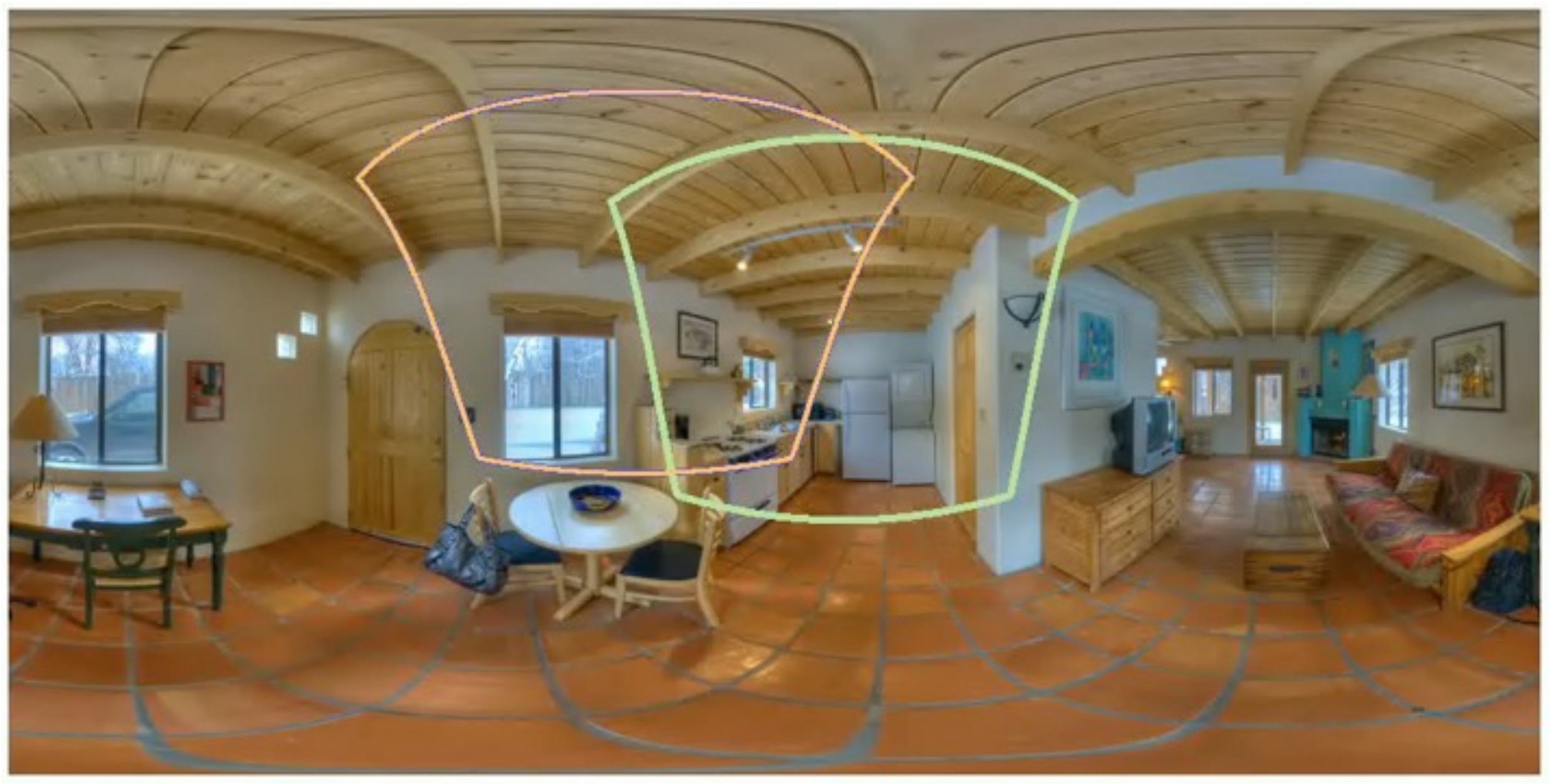}
  \caption{Easy}
  \label{fig:diff_easy}
\end{subfigure}
\hfill
\begin{subfigure}[b]{0.3\textwidth}
  \centering
  \includegraphics[width=\textwidth]{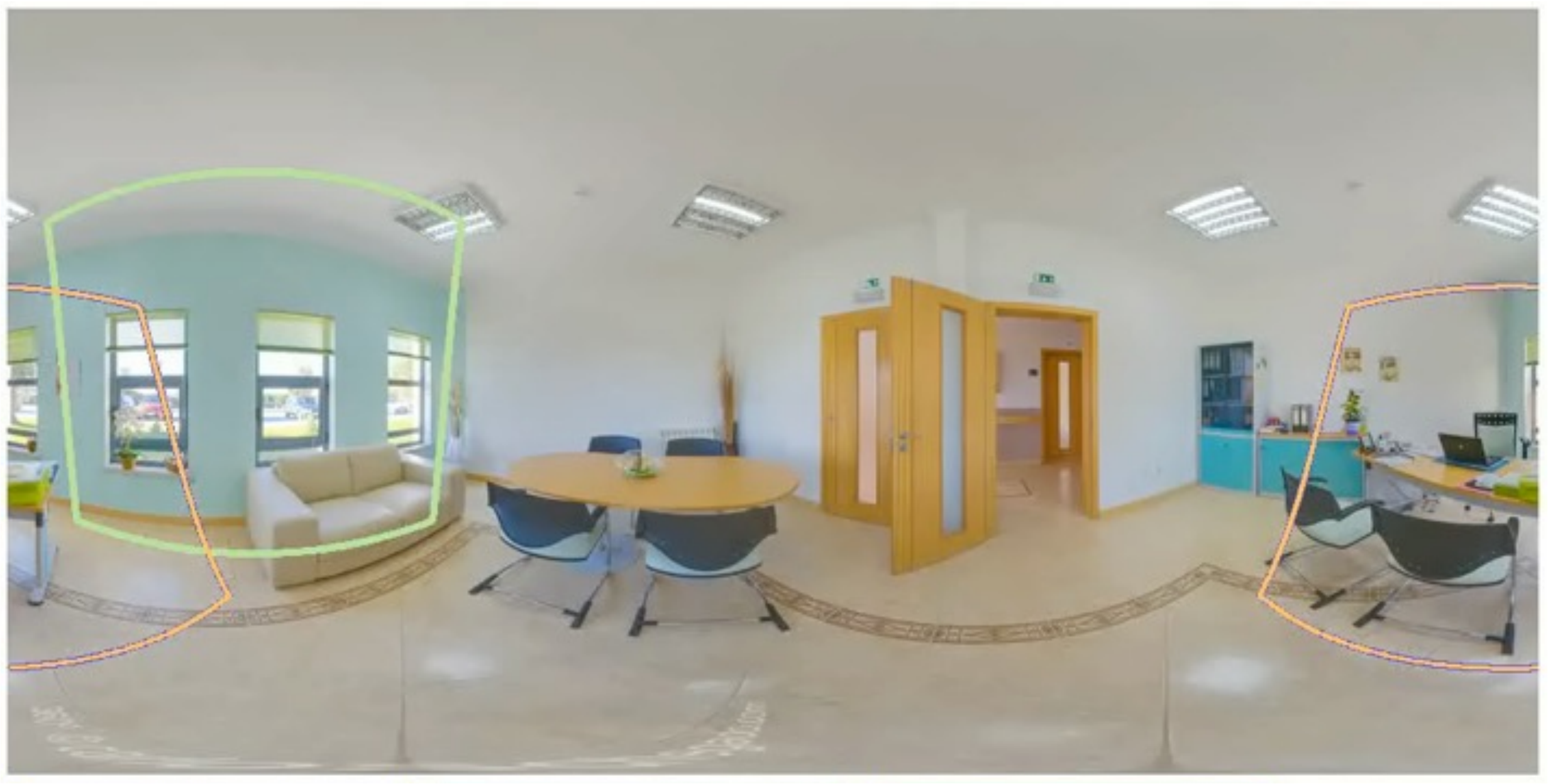}
  \caption{Medium}
  \label{fig:diff_medium}
\end{subfigure}
\hfill
\begin{subfigure}[b]{0.3\textwidth}
  \centering
  \includegraphics[width=\textwidth]{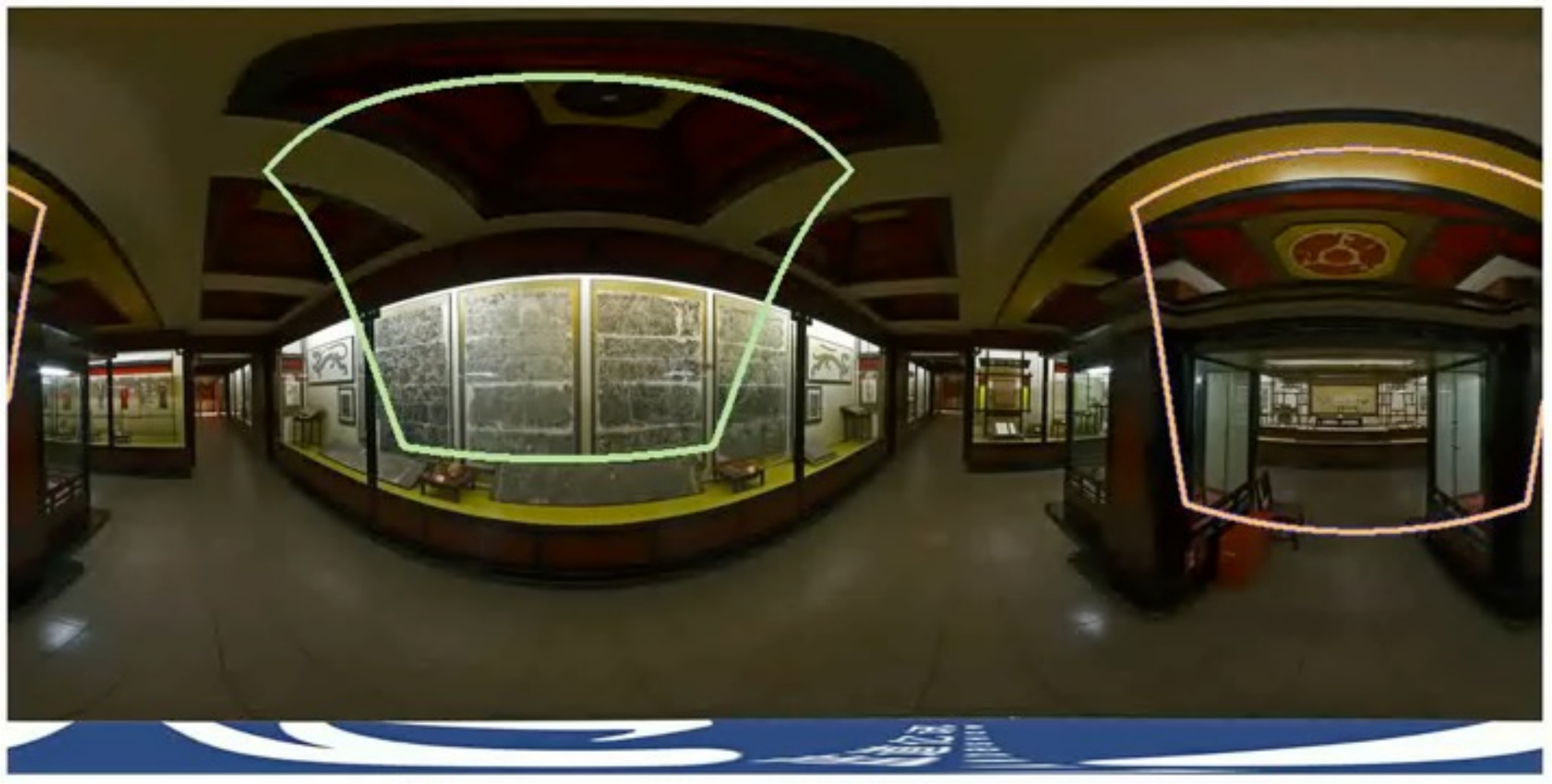}
  \caption{Hard}
  \label{fig:diff_hard}
\end{subfigure}
\caption{The three images are samples of the three levels of difficulties based on the initial view location. The perspective views are shown by the colored bounding boxes projected on the equirectangular image. The initial and target views are respectively shown by the \textcolor{color_pers}{orange} and \textcolor{color_target}{green} boxes (best viewed in color). As shown in \cref{fig:diff_easy}, the initial and target views overlap, and a naive feature matching method could easily localize the views. However, as the difficulties increase and overlap decreases, there needs to be a robust strategy of looking around to localize the target view and not becoming stuck in local minima.}
\vspace{-1.2em}  
\label{fig:diff_initial_conditions}
\end{figure}

We introduce the difficulties to our task in three ways:

\noindent\textbf{Initial View Location.}
We measure localization difficulties by how far the initial rotation is from the target rotation.
We make three levels of `easy', `medium', and `hard' as shown in \cref{fig:diff_initial_conditions}.
The basic premise for the levels depends on the amount of overlap between the initial and target views.
As depicted by the figure, `easy' has the most overlap in FoV and is ostensibly solvable using a strategy to increase the overlap.
We provide detailed explanation of how each level is classified in \cref{apdx:initial_condition}.

\noindent\textbf{Field-of-View.}
We have used two different types of FoV parameters $f=90^\circ$ and $f=60^\circ$.
The intuition is that the wider the FoV, the larger possibility of overlapping with the target view while the agents look around the scene.

\noindent\textbf{Natural Corruptions.}
Since the given target image is not always taken in the same manner as by the agent's sensors, we also added 16 variants of natural corruptions to the target image based on the works of \cite{michaelis2019benchmarking}.
The 16 corruptions consist of blurs, noise models, digital corruptions, and weather models.
The details of the corruptions are in \cref{apdx:corruption}.
The levels of severity used in the benchmarks are $1$ and $3$, which we categorized as low severity and high severity.
The natural corruptions are intended to benchmark the robustness of the agents.

\section{Policy Learning Approach}
\label{sec:learned_appoarch}

\subsection{Problem Formulation}


Our problem formulation follows the model of the Markov decision process (MDP) with the addition of partially observable states.
A typical MDP is defined by states $s \in S$, actions $a \in A$, state transition probability distribution $T(s^\prime | s, a)$, and reward $r = R(s, a)$ \cite{SuttonBarto}.
For the FindView task, information for each state is not fully disclosed to the agent since it can only observe the scene it is looking at.
Therefore, every time the agent takes a new action $a$ , it formulates a new observation $o \in \Omega$, depending on the new state $s^\prime$, with the probability of $O(o | s^\prime, a)$.
The goal of the agent is to choose an action at each time step $t = 0, ..., T$ that maximizes its expected sum of future discounted rewards, which are given by
\begin{equation} \label{eq:return}
R_t = \sum^{T}_{t^\prime=t}\gamma^{t^{\prime} - t}r_{t^\prime},
\end{equation}
where $\gamma \in [0, 1]$ is the discount factor which determines the importance of future rewards and $r_{t^\prime}$ is the reward received at step $t^\prime$.
Reinforcement learning addresses this sequential decision task in which the agent receives only limited environmental feedback, which we employ to train our policy learning approach.
Relating it back to our problem formulation, the agent should choose where to look next based on the observations beforehand and learn a policy $\pi(a_t | o_t)$ that maximizes the expected future discounted reward.

\subsection{Proposed Model}
\label{sec:model}

The model is made of three components:

\noindent\textbf{Encoder Module.}
This module is a network that consists of three convolution layers and a single fully-connected layer to encode the observation $o_t$ into a feature vector $o'_t$.
The encoder module produces $o'_t$ given the network that is parameterized by $\theta$ and is represented by $o'_t = \text{ENC}_{\theta}(o_t)$.

\noindent\textbf{Recurrent Module.}
This module is a recurrent neural network that produces a feature vector $h_t$ based on the previous hidden state vector $h_{t-1}$ and $o'_t$.
The recurrent module is parameterized by $\theta$ and is given by $h_t = \text{GRU}_{\theta}(o'_t, h_{t-1})$.

\noindent\textbf{Actor-Critic Module.}
Finally, in the actor-critic module, policy network $\pi_{\theta}(h_t)$ outputs action distribution of dimension $5$.
The agent then samples an action $a_t$ from the distribution.
Since we train the agent by the policy gradient method \cite{policymethod}, we also use the $h_t$ to estimate the value $v_t$ given by $v_t = V_{\theta}(h_t)$.

Note that each parameter $\theta$ for the networks is different, and we have reused the same notion to point out that the parameters of the networks can be trainable.
Throughout all of the experiments, the feature vector size for $h_t$ and $o'_t$ is $512$.

\subsection{Reward Function Formulation and Reward Shaping}
\label{sec:reward_function}


\noindent\textbf{Success Reward.}
The success reward is given by the equations below,
\begin{equation} \label{eq:success_rewards}
r_{\text{success}} = 
\begin{cases}
    \frac{\alpha}{\| \mathbf{R}_{\text{target}} - \mathbf{R}_t \|_{1} + \beta},& \text{when stopped}\\
    -\alpha,& \text{terminated w/o stopping}\\
    0, & \text{while looking}
\end{cases},
\end{equation}
where $\alpha$, and $\beta$ are scalar values.
From experience, giving large rewards at successful localization makes the training unstable.
This reward function motivates agents to \texttt{stop} frequently and optimizes the trajectories for the peak, which is at $\| \mathbf{R}_{\text{target}} - \mathbf{R}_t \|_{1}=0$.
We found that $\alpha=100$ and $\beta=10$ can enable stable training of agents, and have used these values to train all of the agents.

\noindent\textbf{Warmer-Colder Reward.}
For this reward, if the agent moves closer to the target, the agent would receive a positive reward, while moving farther results in a negative reward \cite{savva2019habitat}.
On step $t$, the agent's \textit{Warmer-Colder} reward is given by,
\begin{equation} \label{eq:hot_or_cold}
r_{\text{dist}} = \gamma_{\text{dist}}(\| \mathbf{R}_{\text{target}} - \mathbf{R}_{t-1} \|_{1} - \| \mathbf{R}_{\text{target}} - \mathbf{R}_t \|_{1}),
\end{equation}
where $\gamma_{\text{dist}}$ is a parameter and we use $0.1$ for all of the training.
This reward motivates the model to understand how their actions affect the relationship between their observations and the target image and motivates agents to discriminate between observations and their distances in respect to the target image.

\noindent\textbf{Slack Reward.}
In order to motivate the agent to take shorter trajectories, we penalize the agents by giving a small negative reward of $r_{\text{slack}}=-0.01$.

Finally, the resulting reward $r_t$ for step $t$ is given as $r_{t} = r_{\text{success}} + r_{\text{dist}} + r_{\text{slack}}$.

\subsection{Training Method}
\label{sec:training}


We trained the policy using the `train' split of the dataset.
Instead of creating a limited number of samples from the equirectangular image, we randomly sampled initial and target views based on their difficulties \cref{sec:findview_dataset}.
We believe that the sampling approach of randomly creating a vast amount of trainable episodes will result in a generalized policy for the task.
With every $n_{\text{update}}$-th update, the policy is evaluated on the `validation' split consistent for all training, and the network weights are saved.
Upon finishing the $N_{\text{update}}$-th update, the weights that have the best metrics throughout the training are then evaluated with the test set (the evaluation results are shown in \cref{sec:experiments}).

During our experiments, we found that training was more stable when the agents were gradually introduced to more difficult episodes after enough training with easier episodes.
Since curriculum learning has been beneficial in reinforcement learning tasks \cite{narvekar2020curriculum} we have employed a difficulty scheduler for this purpose.
We updated the difficulty of the environment for each $n_{\text{difficulty}}$ update.
For all of the experiments, we used a difficulty scheduler to increase the level of difficulty for the initial conditions.
Note that instead of locking a specific difficulty, the scheduler would sample easier difficulties to motivate generalization for the agent.

All of the polices are trained with proximal policy optimization (PPO) \cite{PPO}.
We deployed $N_{\text{envs}}=16$ environments in parallel and trained the policy for $N_{\text{updates}}=30,000$ updates (a total of $61$ million steps).
Validation occurs every $n_{\text{update}}=1,000$ update.
Difficulty for the initial condition levels up for every $n_{\text{difficulty}}=N_{\text{updates}}/3$ updates.
The hyper-parameter settings are shown in \cref{apdx:hyper_params}.

\section{Rule-based Approach}
\label{sec:rulebased_approach}


We consider agents that use feature matching to localize the target view in a heuristic manner.
Our intuition is that local feature descriptors used in applications such as visual SLAM and SfM are robust enough to solve this task.
In short, the agent would detect keypoints and their descriptors for the observation images and perform feature matching.
The matched features are then used to calculate the relative displacements of the view and the action the agent needs to perform to reduce the displacements.
Since observations (perspective and target images) might not have many overlaps, we add distance threshold $d_{\text{thresh}}$ so that low quality matches are not taken into consideration.
During observations with low keypoints or matches, the agent is programmed to look in either \texttt{left} or \texttt{right} and continue looking in the same direction.
See \cref{apdx:pseudo_code} for more detailed explanation and pseudo-codes for the algorithm.

For our experiments, we evaluate two different kinds of feature detectors: SIFT and ORB \cite{lowe2004distinctive,rublee2011orb}.
The values of $d_{\text{thresh}}$ are different for each of the feature detectors and depend on the initial conditions for the task.
We have use the validation set to perform a parameter search for the ideal values of $d_{\text{thresh}}$ for the benchmarks.
The detailed results of the parameter search are shown in \cref{apdx:param_search}.

\begin{table}[t]

\tiny

\setlength\tabcolsep{3pt}
\caption{Benchmarks on the SUN360 Indoor dataset.}
\centering
\begin{tabular}{ c V{2.5} c | c | c | c V{2.5} c | c | c | c V{2.5} c | c | c | c }
\hline\hline
\multirow{2}{*}{Agents}
& \multicolumn{4}{c V{2.5}}{Easy} & \multicolumn{4}{c V{2.5}}{Medium} & \multicolumn{4}{c}{Hard} \T \\ \cline{2-13}
& \multicolumn{1}{c|}{$\varepsilon\downarrow$} & \multicolumn{1}{c|}{$\omega_{\text{stop}}\uparrow$} & \multicolumn{1}{c|}{$\omega_{\text{perf}}\uparrow$} & \multicolumn{1}{cV{2.5}}{$\eta\uparrow$} & \multicolumn{1}{c|}{$\varepsilon\downarrow$} & \multicolumn{1}{c|}{$\omega_{\text{stop}}\uparrow$} & \multicolumn{1}{c|}{$\omega_{\text{perf}}\uparrow$} & \multicolumn{1}{cV{2.5}}{$\eta\uparrow$} & \multicolumn{1}{c|}{$\varepsilon\downarrow$} & \multicolumn{1}{c|}{$\omega_{\text{stop}}\uparrow$} & \multicolumn{1}{c|}{$\omega_{\text{perf}}\uparrow$} & \multicolumn{1}{c}{$\eta\uparrow$} \T \\ \hline
\clineB{0-9}{1}
\texttt{orb}  & 3.41        & 98.7\%       & 89.8\%        &\textbf{87.5\%}& 12.8        & 96.1\%        & 83.6\%        & 68.9\%        & 18.8        & 97.7\%        & 34.3\%        & 28.6\%          \T \\
\texttt{sift} & 3.93        & 96.4\%       & 95.3\%        & 79.9\%        & 4.62        & 98.0\%        & 95.4\%        & 80.0\%        & 9.76        & 97.1\%        & 92.1\%        &\textbf{76.7\%}  \T \\
\hline
\texttt{half}  & 1.14        & 99.6\%       & 93.5\%        & 78.0\%        & 2.41        &\textbf{99.7\%}& 94.8\%        & 72.0\%        & 5.85        &\textbf{99.4\%}& 92.8\%        & 69.1\%          \T \\
\texttt{full}  &\textbf{0.36}&\textbf{100\%}&\textbf{97.6\%}& 84.7\%        &\textbf{0.76}& 99.4\%        &\textbf{97.7\%}&\textbf{80.3\%}&\textbf{2.08}& 99.0\%        &\textbf{97.0\%}&\textbf{76.7\%}    \T \\
\hline\hline
\end{tabular}
\label{table:result_sun360_alpha_indoor}

\setlength\tabcolsep{3pt}
\caption{Benchmarks on the SUN360 Outdoor dataset.}
\centering
\begin{tabular}{ c V{2.5} c | c | c | c V{2.5} c | c | c | c V{2.5} c | c | c | c }
\hline\hline
\multirow{2}{*}{Agents}
& \multicolumn{4}{c V{2.5}}{Easy} & \multicolumn{4}{c V{2.5}}{Medium} & \multicolumn{4}{c}{Hard} \T \\ \cline{2-13}
& \multicolumn{1}{c|}{$\varepsilon\downarrow$} & \multicolumn{1}{c|}{$\omega_{\text{stop}}\uparrow$} & \multicolumn{1}{c|}{$\omega_{\text{perf}}\uparrow$} & \multicolumn{1}{cV{2.5}}{$\eta\uparrow$} & \multicolumn{1}{c|}{$\varepsilon\downarrow$} & \multicolumn{1}{c|}{$\omega_{\text{stop}}\uparrow$} & \multicolumn{1}{c|}{$\omega_{\text{perf}}\uparrow$} & \multicolumn{1}{cV{2.5}}{$\eta\uparrow$} & \multicolumn{1}{c|}{$\varepsilon\downarrow$} & \multicolumn{1}{c|}{$\omega_{\text{stop}}\uparrow$} & \multicolumn{1}{c|}{$\omega_{\text{perf}}\uparrow$} & \multicolumn{1}{c}{$\eta\uparrow$} \T \\ \hline
\clineB{0-9}{1}
\texttt{orb}  & 1.58        & 98.8\%       & 90.7\%          & 88.9\%        & 12.6        & 95.9\%        & 66.4\%        & 48.2\%        & 21.4        & 92.8\%       & 76.2\%        & 63.6\%          \T \\
\texttt{sift} &\textbf{0.33}& 99.7\%       & \textbf{99.7\%} &\textbf{99.7\%}& 3.27        & 99.1\%        &\textbf{97.1\%}&\textbf{87.7\%}& 8.35        & 93.6\%       & 93.0\%        & 76.9\%          \T \\
\hline
\texttt{Indoor} & 1.25        &\textbf{100\%}& 86.7\%          & 75.0\%        & 4.01        & 99.1\%        & 86.4\%        & 69.3\%        & 6.13        & 99.1\%       & 87.0\%        & 67.1\%          \T \\
\texttt{half} & 1.06        & 99.7\%       & 95.9\%          & 87.1\%        &\textbf{1.26}&\textbf{99.7\%}& 95.4\%        & 81.4\%        & 5.49        & 98.8\%       & 92.5\%        & 75.3\%          \T \\
\texttt{full} & 0.79        & 99.7\%       & 96.2\%          & 88.9\%        & 1.53        &\textbf{99.7\%}& 94.2\%        & 81.5\%        &\textbf{3.29}&\textbf{100\%}&\textbf{94.5\%}&\textbf{78.8\%}  \T \\
\hline\hline
\end{tabular}
\label{table:result_sun360_alpha_outdoor}

\setlength\tabcolsep{3pt}
\caption{Benchmarks on the SUN360 Indoor dataset with narrower FoV ($f=60^\circ$).}
\centering
\begin{tabular}{ c V{2.5} c | c | c | c V{2.5} c | c | c | c V{2.5} c | c | c | c }
\hline\hline
\multirow{2}{*}{Agents}
& \multicolumn{4}{c V{2.5}}{Easy} & \multicolumn{4}{c V{2.5}}{Medium} & \multicolumn{4}{c}{Hard} \T \\ \cline{2-13}
& \multicolumn{1}{c|}{$\varepsilon\downarrow$} & \multicolumn{1}{c|}{$\omega_{\text{stop}}\uparrow$} & \multicolumn{1}{c|}{$\omega_{\text{perf}}\uparrow$} & \multicolumn{1}{cV{2.5}}{$\eta\uparrow$} & \multicolumn{1}{c|}{$\varepsilon\downarrow$} & \multicolumn{1}{c|}{$\omega_{\text{stop}}\uparrow$} & \multicolumn{1}{c|}{$\omega_{\text{perf}}\uparrow$} & \multicolumn{1}{cV{2.5}}{$\eta\uparrow$} & \multicolumn{1}{c|}{$\varepsilon\downarrow$} & \multicolumn{1}{c|}{$\omega_{\text{stop}}\uparrow$} & \multicolumn{1}{c|}{$\omega_{\text{perf}}\uparrow$} & \multicolumn{1}{c}{$\eta\uparrow$} \T \\ \hline
\clineB{0-9}{1}
\texttt{orb}  & 16.4        & 85.3\%       & 28.3\%        & 75.2\%        & 37.3        & 72.1\%        & 21.5\%        & 47.2\%        & 51.3        & 71.9\%        & 7.5\%         & 38.2\%         \T \\
\texttt{sift} & 3.42        & 95.2\%       &\textbf{95.2\%}&\textbf{91.8\%}& 9.21        & 89.5\%        & 89.3\%        &\textbf{75.2\%}& 23.9        & 85.8\%        &\textbf{85.2\%}&\textbf{66.8\%} \T \\
\hline
\texttt{full} &\textbf{2.45}&\textbf{99.9\%}& 92.9\%       & 80.1\%        &\textbf{6.16}&\textbf{99.3\%}&\textbf{90.7\%}& 67.2\%        &\textbf{14.2}&\textbf{99.6\%}&\textbf{85.2\%}& 59.9\%         \T \\
\hline\hline
\end{tabular}
\label{table:result_sun360_gamma_indoor}

\setlength\tabcolsep{3pt}
\caption{Benchmarks on the SUN360 Outdoor with narrower FoV ($f=60^\circ$).}
\centering
\begin{tabular}{ c V{2.5} c | c | c | c V{2.5} c | c | c | c V{2.5} c | c | c | c }
\hline\hline
\multirow{2}{*}{Agents}
& \multicolumn{4}{c V{2.5}}{Easy} & \multicolumn{4}{c V{2.5}}{Medium} & \multicolumn{4}{c}{Hard} \T \\ \cline{2-13}
& \multicolumn{1}{c|}{$\varepsilon\downarrow$} & \multicolumn{1}{c|}{$\omega_{\text{stop}}\uparrow$} & \multicolumn{1}{c|}{$\omega_{\text{perf}}\uparrow$} & \multicolumn{1}{cV{2.5}}{$\eta\uparrow$} & \multicolumn{1}{c|}{$\varepsilon\downarrow$} & \multicolumn{1}{c|}{$\omega_{\text{stop}}\uparrow$} & \multicolumn{1}{c|}{$\omega_{\text{perf}}\uparrow$} & \multicolumn{1}{cV{2.5}}{$\eta\uparrow$} & \multicolumn{1}{c|}{$\varepsilon\downarrow$} & \multicolumn{1}{c|}{$\omega_{\text{stop}}\uparrow$} & \multicolumn{1}{c|}{$\omega_{\text{perf}}\uparrow$} & \multicolumn{1}{c}{$\eta\uparrow$} \T \\ \hline
\clineB{0-9}{1}
\texttt{orb}  & 15.0        & 85.3\%       & 25.9\%        & 76.8\%        & 28.9        & 72.3\%        & 13.8\%        & 44.7\%        & 51.6        & 67.1\%        & 17.0\%        & 47.7\%          \T \\
\texttt{sift} & 1.79        & 98.0\%       &\textbf{97.7\%}&\textbf{97.3\%}& 10.7        & 88.2\%        & 87.9\%        &\textbf{80.5\%}& 22.8        & 84.4\%        & 84.1\%        & 66.5\%          \T \\
\hline
\texttt{full} &\textbf{1.25}&\textbf{99.7\%}& 93.9\%       & 89.3\%        &\textbf{4.92}&\textbf{99.7\%}&\textbf{91.9\%}& 79.3\%        &\textbf{7.36}&\textbf{90.2\%}&\textbf{90.2\%}&\textbf{72.6\%}  \T \\
\hline\hline
\end{tabular}
\label{table:result_sun360_gamma_outdoor}
\vspace{-1.2em}  

\end{table}

\begin{table}[t]

\tiny

\setlength\tabcolsep{3pt}
\caption{Benchmarks on the 360-Indoor dataset.}
\centering
\begin{tabular}{ c V{2.5} c | c | c | c V{2.5} c | c | c | c V{2.5} c | c | c | c }
\hline\hline
\multirow{2}{*}{Agents}
& \multicolumn{4}{c V{2.5}}{Easy} & \multicolumn{4}{c V{2.5}}{Medium} & \multicolumn{4}{c}{Hard} \T \\ \cline{2-13}
& \multicolumn{1}{c|}{$\varepsilon\downarrow$} & \multicolumn{1}{c|}{$\omega_{\text{stop}}\uparrow$} & \multicolumn{1}{c|}{$\omega_{\text{perf}}\uparrow$} & \multicolumn{1}{cV{2.5}}{$\eta\uparrow$} & \multicolumn{1}{c|}{$\varepsilon\downarrow$} & \multicolumn{1}{c|}{$\omega_{\text{stop}}\uparrow$} & \multicolumn{1}{c|}{$\omega_{\text{perf}}\uparrow$} & \multicolumn{1}{cV{2.5}}{$\eta\uparrow$} & \multicolumn{1}{c|}{$\varepsilon\downarrow$} & \multicolumn{1}{c|}{$\omega_{\text{stop}}\uparrow$} & \multicolumn{1}{c|}{$\omega_{\text{perf}}\uparrow$} & \multicolumn{1}{c}{$\eta\uparrow$} \T \\ \hline
\clineB{0-9}{1}
\texttt{orb}  & 4.71        & 97.9\%       & 75.7\%        & 89.9\%        & 12.6        & 94.3\%        & 70.0\%        & 64.3\%        & 25.4        & 94.9\%        & 36.9\%        & 50.5\%          \T \\
\texttt{sift} & 4.65        & 97.3\%       & 93.1\%        & 90.4\%        & 11.0        & 93.1\%        & 88.0\%        & 64.3\%        & 17.2        & 91.0\%        & 87.1\%        & 70.4\%          \T \\
\hline
\texttt{sun360} & 0.94        & 99.7\%       & 91.9\%        & 86.2\%        & 3.66        & 99.4\%        & 89.5\%        & 81.1\%        & 4.58        & 98.8\%        & 88.6\%        & 75.7\%          \T \\
\texttt{fine-tuned} &\textbf{0.86}&\textbf{100\%}&\textbf{97.9\%}&\textbf{91.3\%}&\textbf{1.80}&\textbf{99.7\%}&\textbf{96.7\%}&\textbf{85.8\%}&\textbf{3.42}&\textbf{99.4\%}&\textbf{95.6\%}&\textbf{77.6\%}  \T \\
\hline\hline
\end{tabular}
\label{table:result_360Indoor_alpha}

\setlength\tabcolsep{3pt}
\caption{Benchmarks on the 360-Indoor dataset with narrower FoV ($f=60^\circ$).}
\centering
\begin{tabular}{ c V{2.5} c | c | c | c V{2.5} c | c | c | c V{2.5} c | c | c | c }
\hline\hline
\multirow{2}{*}{Agents}
& \multicolumn{4}{c V{2.5}}{Easy} & \multicolumn{4}{c V{2.5}}{Medium} & \multicolumn{4}{c}{Hard} \T \\ \cline{2-13}
& \multicolumn{1}{c|}{$\varepsilon\downarrow$} & \multicolumn{1}{c|}{$\omega_{\text{stop}}\uparrow$} & \multicolumn{1}{c|}{$\omega_{\text{perf}}\uparrow$} & \multicolumn{1}{cV{2.5}}{$\eta\uparrow$} & \multicolumn{1}{c|}{$\varepsilon\downarrow$} & \multicolumn{1}{c|}{$\omega_{\text{stop}}\uparrow$} & \multicolumn{1}{c|}{$\omega_{\text{perf}}\uparrow$} & \multicolumn{1}{cV{2.5}}{$\eta\uparrow$} & \multicolumn{1}{c|}{$\varepsilon\downarrow$} & \multicolumn{1}{c|}{$\omega_{\text{stop}}\uparrow$} & \multicolumn{1}{c|}{$\omega_{\text{perf}}\uparrow$} & \multicolumn{1}{c}{$\eta\uparrow$} \T \\ \hline
\clineB{0-9}{1}
\texttt{orb}  & 17.8        & 80.2\%        & 18.9\%        & 65.1\%        & 30.7        & 68.8\%        & 15.9\%        & 40.5\%        & 53.3        & 69.1\%        & 10.5\%        & 35.1\% \T \\
\texttt{sift} & 6.31        & 91.6\%        &\textbf{90.1\%}&\textbf{88.1\%}& 12.5        & 83.5\%        & 82.9\%        & 62.7\%        & 31.0        & 78.7\%        &\textbf{78.4\%}& \textbf{62.1\%} \T \\
\hline
\texttt{SUN360} & 5.59        & 99.1\%        & 83.2\%        & 79.6\%        & 13.5        & 97.3\%        & 77.8\%        & 65.9\%        &\textbf{18.2}& 96.4\%        &\textbf{78.4\%}& 57.6\% \T \\
\texttt{fine-tuned} &\textbf{4.52}&\textbf{99.4\%}& 83.5\%        & 83.0\%        &\textbf{12.1}&\textbf{98.8\%}&\textbf{83.8\%}&\textbf{68.5\%}& 22.5        &\textbf{97.3\%}& 78.1\%        & 55.4\% \T \\
\hline\hline
\end{tabular}
\label{table:result_360Indoor_gamma}

\vspace{-1.2em}  

\end{table}

\begin{table}[t]

\tiny

\setlength\tabcolsep{3pt}
\caption{Benchmarks on the SUN360 Indoor dataset with low corruption.}
\centering
\begin{tabular}{ c || c V{2.5} c | c | c | c V{2.5} c | c | c | c V{2.5} c | c | c | c V{2.5} c | c | c | c }
\hline\hline
\multirow{2}{*}{Agents} & \multirow{2}{*}{Clear} & \multicolumn{4}{c V{2.5}}{Blur} & \multicolumn{4}{c V{2.5}}{Noise} & \multicolumn{4}{c V{2.5}}{Digital} & \multicolumn{4}{c}{Weather} \T \\ \cline{3-18}
& & \multicolumn{1}{c|}{Moti.} & \multicolumn{1}{c|}{Defo.} & \multicolumn{1}{c|}{Glass} & \multicolumn{1}{cV{2.5}}{Gauss.} & \multicolumn{1}{c|}{Gauss.} & \multicolumn{1}{c|}{Impul.} & \multicolumn{1}{c|}{Shot} & \multicolumn{1}{cV{2.5}}{Spec.} & \multicolumn{1}{c|}{Bright.} & \multicolumn{1}{c|}{Contr.} & \multicolumn{1}{c|}{Satur.} & \multicolumn{1}{cV{2.5}}{JPEG} & \multicolumn{1}{c|}{Snow} & \multicolumn{1}{c|}{Spat.} & \multicolumn{1}{c|}{Fog} & \multicolumn{1}{c}{Frost} \T \\ \hline
\clineB{0-17}{1}
\texttt{orb}  & 17.9 & 19.5 & 24.4 & 21.9 & 16.3 & 16.1 & 18.7 & 20.1 & 17.9 & 19.5 & 30.1 & 23.4 & 18.9 & \textbf{25.0} & 21.7 & \textbf{54.5} & \textbf{30.4} \T \\
\texttt{sift}  & 9.73 & 46.1 & 51.9 & 49.4 & 30.6 & 33.6 & 41.4 & 41.2 & 40.3 & \textbf{8.57} & \textbf{19.6} & 21.1 & 29.6 & 76.5 & 9.72 & 75.2 & 106.3 \T \\
\hline
\texttt{full} & \textbf{0.02} & 5.82 & 2.08 & \textbf{0.35} & \textbf{0.00} & \textbf{0.85} & \textbf{0.12} & \textbf{0.05} & \textbf{0.02} & 90.6 & 107.7 & 80.7 & 1.77 & 91.4 & \textbf{0.02} & 112.2 & 94.9 \T \\
\texttt{fine-tuned} & 0.07 & \textbf{2.50} & \textbf{1.23} & 0.38 & 1.95 & 2.13 & 2.17 & 1.70 & 3.70 & 12.9 & 66.0 & \textbf{5.23} & \textbf{0.12} & 29.9 & 3.75 & 92.2 & 31.05 \T \\
\hline\hline
\end{tabular}
\label{table:result_corrupted_indoor_low}

\setlength\tabcolsep{3pt}
\caption{Benchmarks on the SUN360 Indoor dataset with high corruption.}
\centering
\begin{tabular}{ c || c V{2.5} c | c | c | c V{2.5} c | c | c | c V{2.5} c | c | c | c V{2.5} c | c | c | c }
\hline\hline
\multirow{2}{*}{Agents} & \multirow{2}{*}{Clear} & \multicolumn{4}{c V{2.5}}{Blur} & \multicolumn{4}{c V{2.5}}{Noise} & \multicolumn{4}{c V{2.5}}{Digital} & \multicolumn{4}{c}{Weather} \T \\ \cline{3-18}
& & \multicolumn{1}{c|}{Moti.} & \multicolumn{1}{c|}{Defo.} & \multicolumn{1}{c|}{Glass} & \multicolumn{1}{cV{2.5}}{Gauss.} & \multicolumn{1}{c|}{Gauss.} & \multicolumn{1}{c|}{Impul.} & \multicolumn{1}{c|}{Shot} & \multicolumn{1}{cV{2.5}}{Spec.} & \multicolumn{1}{c|}{Bright.} & \multicolumn{1}{c|}{Contr.} & \multicolumn{1}{c|}{Satur.} & \multicolumn{1}{cV{2.5}}{JPEG} & \multicolumn{1}{c|}{Snow} & \multicolumn{1}{c|}{Spat.} & \multicolumn{1}{c|}{Fog} & \multicolumn{1}{c}{Frost} \T \\ \hline
\clineB{0-17}{1}
\texttt{orb} & 17.9 & 31.5 & 45.7 & 25.5 & 45.9 & 29.2 & 27.5 & 18.6 & 23.7 & \textbf{17.9} & \textbf{68.3} & \textbf{20.6} & 16.7 & 57.0 & 38.8 & 97.9 & 84.0 \T \\
\texttt{sift} & 9.73 & 88.5 & 85.3 & 91.1 & 79.8 & 87.3 & 79.7 & 82.5 & 80.2 & 37.4 & 89.0 & 27.9 & 37.9 & 94.7 & 104.2 & 103.4 & 110.7 \T \\
\hline
\texttt{full} & \textbf{0.02} & 6.87 & \textbf{3.00} & 4.77 & \textbf{3.05} & 25.4 & 28.6 & 10.6 & 5.52 & 80.9 & 114.9 & 95.2 & \textbf{0.02} & 108.3 & 8.38 & 112.5 & 106.6 \T \\
\texttt{fine-tuned} & 0.07 & \textbf{3.35} & 4.13 & \textbf{3.80} & 3.45 & \textbf{6.17} & \textbf{5.45} & \textbf{4.98} & \textbf{3.73} & 37.8 & 98.9 & 34.1 & 4.43 & \textbf{52.0} & \textbf{6.10} & \textbf{94.0} & \textbf{53.8} \T \\
\hline\hline
\end{tabular}
\label{table:result_corrupted_indoor_high}

\setlength\tabcolsep{3pt}
\caption{Benchmarks on the SUN360 Outdoor dataset with low corruption.}
\centering
\begin{tabular}{ c || c V{2.5} c | c | c | c V{2.5} c | c | c | c V{2.5} c | c | c | c V{2.5} c | c | c | c }
\hline\hline
\multirow{2}{*}{Agents} & \multirow{2}{*}{Clear} & \multicolumn{4}{c V{2.5}}{Blur} & \multicolumn{4}{c V{2.5}}{Noise} & \multicolumn{4}{c V{2.5}}{Digital} & \multicolumn{4}{c}{Weather} \T \\ \cline{3-18}
& & \multicolumn{1}{c|}{Moti.} & \multicolumn{1}{c|}{Defo.} & \multicolumn{1}{c|}{Glass} & \multicolumn{1}{cV{2.5}}{Gauss.} & \multicolumn{1}{c|}{Gauss.} & \multicolumn{1}{c|}{Impul.} & \multicolumn{1}{c|}{Shot} & \multicolumn{1}{cV{2.5}}{Spec.} & \multicolumn{1}{c|}{Bright.} & \multicolumn{1}{c|}{Contr.} & \multicolumn{1}{c|}{Satur.} & \multicolumn{1}{cV{2.5}}{JPEG} & \multicolumn{1}{c|}{Snow} & \multicolumn{1}{c|}{Spat.} & \multicolumn{1}{c|}{Fog} & \multicolumn{1}{c}{Frost} \T \\ \hline
\clineB{0-17}{1}
\texttt{orb}  & 13.6 & 11.9 & 13.9 & 10.8 & 16.3 & 9.33 & 10.6 & 12.9 & 5.65 & 9.83 & \textbf{17.8} & 17.1 & 14.7 & \textbf{18.7} & 12.2 & \textbf{42.8} & \textbf{20.7} \T \\
\texttt{sift} & 0.28 & 66.7 & 69.4 & 68.8 & 47.4 & 65.8 & 72.0 & 70.8 & 58.8 & 9.63 & 21.3 & 31.6 & 45.1 & 82.6 & 10.8 & 84.9 & 97.7 \T \\
\hline
\texttt{full} & \textbf{0.02} & \textbf{0.87} & \textbf{0.12} & \textbf{0.63} & \textbf{0.10} & \textbf{0.12} & \textbf{0.07} & \textbf{0.10} & \textbf{0.15} & 67.7 & 109.9 & 38.7 & \textbf{0.03} & 86.2 & \textbf{0.00} & 118.7 & 90.8 \T \\
\texttt{fine-tuned} & 5.40 & 6.62 & 2.40 & 5.28 & 3.27 & 3.00 & 4.40 & 1.18 & 0.95 & \textbf{4.82} & 72.42 & \textbf{9.53} & 3.30 & 24.1 & 0.85 & 79.7 & 37.5 \T \\
\hline\hline
\end{tabular}
\label{table:result_corrupted_outdoor_low}

\setlength\tabcolsep{3pt}
\caption{Benchmarks on the SUN360 Outdoor dataset with high corruption.}
\centering
\begin{tabular}{ c || c V{2.5} c | c | c | c V{2.5} c | c | c | c V{2.5} c | c | c | c V{2.5} c | c | c | c }
\hline\hline
\multirow{2}{*}{Agents} & \multirow{2}{*}{Clear} & \multicolumn{4}{c V{2.5}}{Blur} & \multicolumn{4}{c V{2.5}}{Noise} & \multicolumn{4}{c V{2.5}}{Digital} & \multicolumn{4}{c}{Weather} \T \\ \cline{3-18}
& & \multicolumn{1}{c|}{Moti.} & \multicolumn{1}{c|}{Defo.} & \multicolumn{1}{c|}{Glass} & \multicolumn{1}{cV{2.5}}{Gauss.} & \multicolumn{1}{c|}{Gauss.} & \multicolumn{1}{c|}{Impul.} & \multicolumn{1}{c|}{Shot} & \multicolumn{1}{cV{2.5}}{Spec.} & \multicolumn{1}{c|}{Bright.} & \multicolumn{1}{c|}{Contr.} & \multicolumn{1}{c|}{Satur.} & \multicolumn{1}{cV{2.5}}{JPEG} & \multicolumn{1}{c|}{Snow} & \multicolumn{1}{c|}{Spat.} & \multicolumn{1}{c|}{Fog} & \multicolumn{1}{c}{Frost} \T \\ \hline
\clineB{0-17}{1}
\texttt{orb}  & 13.6 & 53.7 & 50.4 & 18.7 & 48.5 & 5.33 & 15.9 & 12.7 & 13.6 & \textbf{4.90} & \textbf{59.2} & \textbf{4.80} & 4.50 & \textbf{31.5} & 23.4 & \textbf{67.8} & \textbf{52.2} \T \\
\texttt{sift} & 0.28 & 87.1 & 86.4 & 86.3 & 84.4 & 89.4 & 88.3 & 87.7 & 92.1 & 67.5 & 84.6 & 47.0 & 61.1 & 90.8 & 81.0 & 97.2 & 99.5 \T \\
\hline
\texttt{full} & \textbf{0.02} & \textbf{2.20} & \textbf{0.18} & \textbf{1.87} & \textbf{0.13} & \textbf{2.72} & \textbf{6.18} & \textbf{1.57} & \textbf{0.50} & 101.1 & 114.0 & 79.1 & \textbf{0.05} & 115.2 & \textbf{0.35} & 102.0 & 98.6 \T \\
\texttt{fine-tuned}  & 5.40 & 8.48 & 4.72 & 10.0 & 8.22 & 4.23 & 7.62 & 7.60 & 4.63 & 41.7 & 87.5 & 26.9 & 2.92 & 63.2 & 1.82 & 83.7 & 56.9 \T \\
\hline\hline
\end{tabular}
\label{table:result_corrupted_outdoor_high}

\vspace{-1.2em}  

\end{table}

\section{Experiments}
\label{sec:experiments}

\noindent\textbf{Datasets.}
As described in \cref{sec:findview_dataset}, we use SUN360 Indoor, SUN360 Outdoor, and 360-Indoor.
We sampled episodes according to the specified difficulties (initial conditions and corruption severities) and used the same test episodes throughout the experiments for each dataset.

\noindent\textbf{Agents.}
We use the agents described in \cref{sec:learned_appoarch} and \cref{sec:rulebased_approach}:
\setlist{nolistsep}
\begin{itemize}[noitemsep]
\item \texttt{orb}: Rule-based agent using ORB feature detector.
\item \texttt{sift}: Rule-based agent using SIFT feature detector.
\item \texttt{half}: PPO agent trained with $N_{\text{updates}}=15,000$.
\item \texttt{full}: PPO agent trained with $N_{\text{updates}}=30,000$.
\end{itemize}
Note that there are experiment-specific agents like \texttt{fine-tuned}, which are PPO agent that is fine-tuned on a different dataset.
Also note that for \texttt{orb} and \texttt{sift}, we have used specific $d_{\text{thresh}}$ values for each \textbf{difficulty} (`easy', `medium', and `hard') and \textbf{datasets} (SUN360's indoor and outdoor, and 360-Indoor), which were found using the validation split of the respective datasets.
On the other hand, PPO agents use the same weights throughout each difficulty.

\noindent\textbf{Evaluations.}
We evaluate each agent in the metrics described in \cref{sec:findview_evaluations}, which are:
\setlist{nolistsep}
\begin{itemize}[noitemsep]
\item $\varepsilon$: Localization error ($\downarrow$ lower is better).
\item $\omega_{\text{stop}}$ ($\%$): Percentage of stops ($\uparrow$ higher is better).
\item $\omega_{\text{stop}}$ ($\%$): Percentage of perfects ($\uparrow$ higher is better).
\item $\eta$ ($\%$): SPL ($\uparrow$ higher is better).
\end{itemize}

Implementation details are provided in \cref{apdx:implementation details}.

\subsection{Benchmarks on the SUN360 Dataset}
\label{sec:sun360_experiment}

For the first experiment, the agents are in an environment where the target image is completely recoverable.
\cref{table:result_sun360_alpha_indoor} shows the results on the SUN360 indoor dataset.
We observe that \texttt{full} achieved the lowest error-rate $\varepsilon$ which outperformed the rule-based method by large margins.
\texttt{half} also achieved $\approx 2$ times better than \texttt{sift}, which seems to be more accurate than \texttt{orb} in most cases.
On `easy', \texttt{full} has stopped on every episode and has localized $97.6\%$ perfectly. 
Trained methods (\texttt{half} and \texttt{full}) came first and second on most metrics, except for SPL $\eta$ in `easy'.
SPL is a metric that measures the success rate weighted by the path length \cite{anderson2018evaluation}, and is therefore influenced by the length of the path it took for perfect localization.
Looking at the values of $\varepsilon$ and $\omega_{\text{perf}}$, the agent localized perfectly more times than the other methods.
This means that even though the number of perfection cases is low, \texttt{orb} achieved higher SPL because it had followed closely with the oracle trajectories.
Trained methods localized with drastically low $\varepsilon$ values, with the trade-off of having slightly inefficient trajectories.
This is a trend seen in the other benchmarks results such as the SUN360 outdoor dataset shown in \cref{table:result_sun360_alpha_outdoor}.
We encourage the readers to see the qualitative results in the supplemented video.

As for the results in \cref{table:result_sun360_alpha_outdoor}, \texttt{sift} achieved the lowest $\varepsilon$ for `easy'.
We stress the fact that values for $d_{thresh}$ are tailored for each difficulty for the rule-based methods.
In practice, the agents would not know whether the next episode is easy or hard.
From this perspective, trained methods are flexible since the network weights are consistent among difficulties.
We also show that \texttt{Indoor}, which is \texttt{full} trained in the indoor dataset, is localizing competitively, signifying its generalization capabilities.

For narrower FoV, as shown in \cref{table:result_sun360_gamma_indoor} and \cref{table:result_sun360_gamma_outdoor}, $\varepsilon$ tends to be worse which indicates the raise in difficulty.
We observe that the trained method (\texttt{full}) still achieved the lowest localization error while managing to attain competitive SPL.  

\subsection{Benchmarks on the 360-Indoor dataset.}
\label{sec:360indoor_experiment}

We evaluated the agents on the 360-Indoor dataset in the same manner as \cref{sec:sun360_experiment}.
In the experiment, we have used \texttt{sun360} that adopts the same weights used for \texttt{full} in \cref{table:result_360Indoor_alpha}.
We have also fine-tuned the weights by training the agent for an additional $N_{\text{updates}}=15,000$ on this dataset, which is denoted by \texttt{fine-tuned}.
Surprisingly, as shown in \cref{table:result_360Indoor_alpha}, \texttt{fine-tuned} achieves the best metrics across the board.
Since 360-Indoor and SUN360 datasets do not share any scenes, we believe that this result comes from the additional training by fine-tuning the network.
Similar pattern can be seen in the narrow FoV setting shown in \cref{table:result_360Indoor_gamma}.


\subsection{Benchmarks with Natural Corruptions}
\label{sec:corruptions}

Finally, we have evaluated the agents by adding natural corruptions to the target images.
We have used the SUN360 Indoor and Outdoor datasets with low and high levels of corruption.
We adopt the corruptions used in \cite{michaelis2019benchmarking,kamann2020benchmarking} and show a sample visualization in \cref{apdx:corruption}.
We averaged the localization errors $\varepsilon$ between all of the difficulties (`easy,' `medium,' and `hard'), and displayed a single value for each corruption.
`Clear' indicates that the result is without any corruption.
Each result is shown in \cref{table:result_corrupted_indoor_low,table:result_corrupted_indoor_high,table:result_corrupted_outdoor_low,table:result_corrupted_outdoor_high}.
\texttt{orb}, \texttt{sift}, and \texttt{full} uses the same parameters and weights as in \cref{sec:sun360_experiment}, while \texttt{fine-tuned} is \texttt{full} that is trained for an additional $N_{\text{updates}}=15,000$ with the corruptions as data augmentation.

Generally, learnt methods \texttt{fine-tune} and \texttt{full} achieved strong results for all blur and noise corruptions as well as `JPEG compression' and `spatter'.
Fine-tuned model performed robustly for both severities while improving `brightness' and `saturation' as well as weather corruptions that were difficult for \texttt{full}.
However, \texttt{fine-tune} appears to lose accuracy in low severities such as `clear'.
We believe this is a trade-off of learning to localize \textit{similar} views.
We hope to improve these metrics for future works by enabling a more discriminative model.
The current model is tailored to recognize exact views as we concatenate the images by early fusion.
We believe that using a Siamese architecture with metric learning loss could improve the model to learn which views are similar.

On the other hand, \texttt{orb} displayed competitive results for the digital and weather domain, which is as expected because it uses hand-crafted features that are robust against noise and changes in lighting.
However, \texttt{fine-tune} achieved better results for most corruptions and we believe that this could be improved with more training. 


\subsection{Runtimes of the Agents}
\label{sec:runtimes}

We timed the runtime in frames-per-second (FPS) of each agent and have presented the results in \cref{table:runtimes}.
This only measures the agent's processing speed and does not include the environment.
Not only has learnt method improved the localization results, it is also faster by $\approx 6$ times compared to \texttt{orb} and $\approx 22$ times compared to \texttt{sift}. 

\begin{table}[t]
\caption{Runtimes of the Agents.}
\centering
\begin{tabular}{c V{2.5} c | c | c}
Agents & \texttt{orb} & \texttt{sift} & \texttt{full} \\
\hline
FPS $\uparrow$   & 120.9 & 33.3 & 733.6 \\
\end{tabular}
\vspace{-1.2em}  
\label{table:runtimes}
\end{table}

\section{Limitations}
\label{sec:limitations}

The limitation of the FindView task is that we assume that the target images and observations are taken from the same 3D position and therefore does not consider camera translations.
Experiments were performed where these target images are corrupted, but still sampled from the same source.
In real-life applications, we would need to extend this task so that agents learn to ``look around" to find similar views and extend the action space to add the functionality for agents to determine that the target image could not be found from the current 3D position.

We have tried to make the training more robust with reward shaping and curriculum learning (\cref{sec:reward_function,sec:training}).
However, due to the difficulties in reinforcement learning, the results can vary using different seeds.
We made sure that the training and results were reproducible using the same seeds.
The training process of the agents used in the benchmarks are described in \cref{apdx:training}.

\section{Potential negative societal impacts}
\label{sec:societal_impacts}

This task can be extended further by allowing the target image to be anything.
For example, instead of an image, we could potentially train the agent to find a specific object or person in a scene by looking around.
This would increase the motivation of searching and tracking people using PTZ like cameras which are quite common for CCTV in public environments.
While the application could increase security and safety, it can be abused by authorities and cause privacy infringement in certain situations.

\section{Conclusion}
\label{sec:conclusions}

We proposed a novel and straightforward task of precise target view localization for look around agents called the FindView task.
This task imitates the view movements of PTZ cameras or user interfaces for $360^\circ$ mediums.
We have introduced two agents for solving this task: rule-based agent and policy learning agent.
The learned agent is highly precise and localizes in heavily corrupt or novel scenes as shown by the extensive evaluations and benchmarks.


\clearpage
\begin{appendices}

\section{Simulator}
\label{apdx:simulator}

\begin{figure}[ht]
\centering
\includegraphics[width=0.4\linewidth]{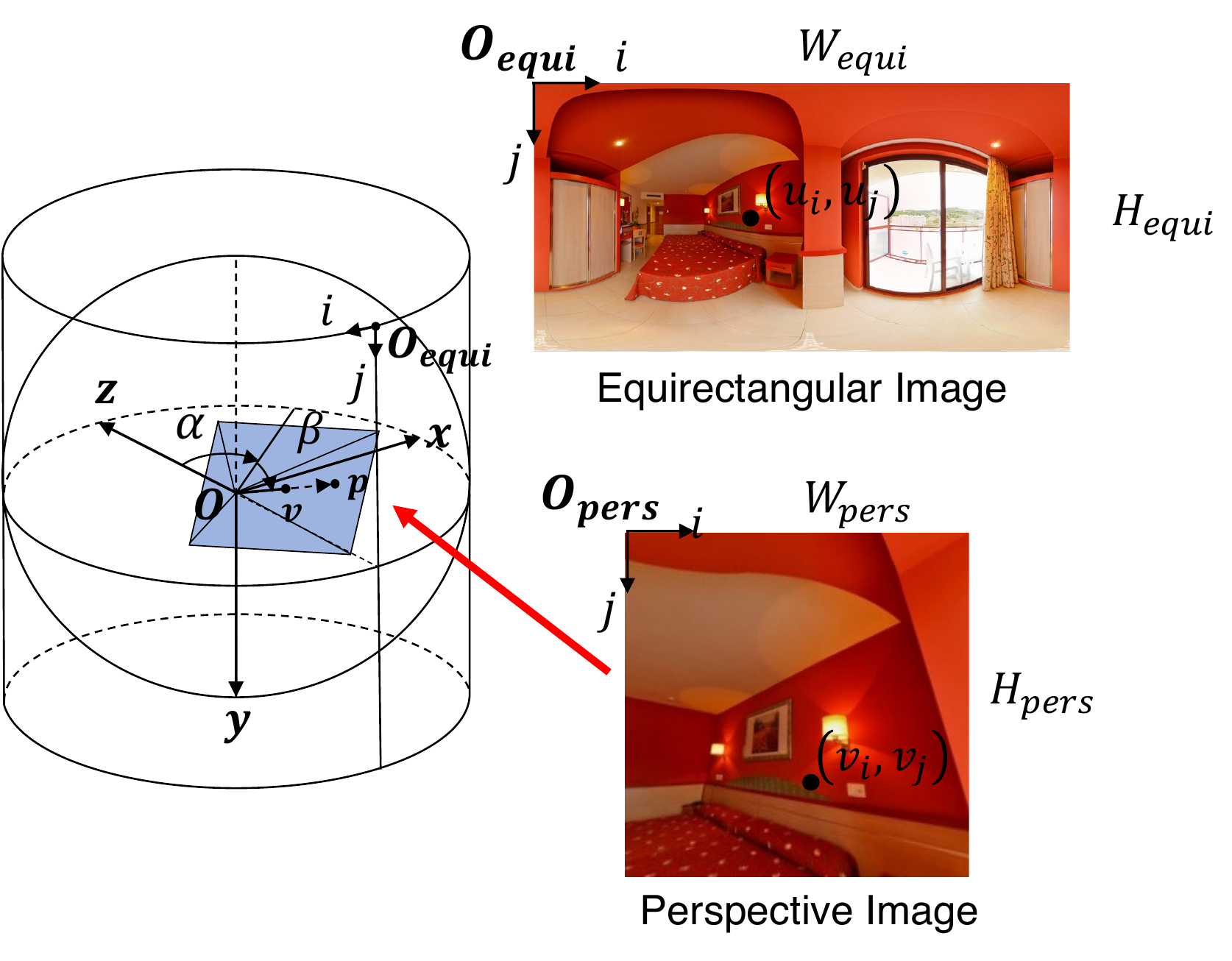}
\caption{The coordinate system for transforming the equirectangular image to a perspective image based on rotation $(\alpha, \beta)$.}
\label{fig:pano2pers}
\end{figure}

The simulator transforms the equirectangular image into a perspective image given the rotation of the vector, which passes through the center of the perspective image.
As shown in \cref{fig:pano2pers}, for the equirectangular image having the dimensions $W_{\text{equi}}$ and $H_{\text{equi}}$, a point $\mathbf{u} = (u_{i}, u_{j})^{\intercal}$ in the equirectangular image coordinates is transformed into a unit vector $\mathbf{p} = (p_{x}, p_{y}, p_{z})^{\intercal}$.

Our goal is to create a perspective image by gathering the corresponding pixels from the equirectangular image.
First, we would need to define the intrinsic parameters for the virtual perspective image.
Given the horizontal field of view $f$ and the dimensions for the perspective image $(W_{\text{pers}}, H_{\text{pers})}$ the intrinsic matrix is calculated as:
\begin{equation} \label{eq:K}
\mathbf{K} = \begin{bmatrix}
    f' & 0 & W_{\text{pers}}/2 \\
    0 & f' & H_{\text{pers}}/2 \\
    0 & 0 & 1
\end{bmatrix},
\end{equation}
where $f' = W_{\text{pers}}/(2 \times \tan((f\times \frac{\pi}{180^\circ}) / 2))$ is the focal length.

Given the rotation angles $\alpha$ and $\beta$ (corresponds to yaw and pitch of the center of the perspective image relative to the $z$-axis), the rotation matrix $\mathbf{R}$ is obtained.

We need to use the intrinsic matrix to find each orientation that maps pixels on the perspective image $\mathbf{v}(v_{i}, v_{j})$ to $\mathbf{p}$ on the equirectangular image, which is shown below:
\begin{equation} \label{eq:cameramatrix}
\begin{bmatrix}
    p_x \\
    p_y \\
    p_z
\end{bmatrix} = \mathbf{R} \mathbf{K}^{-1} \begin{bmatrix}
    v_i \\
    v_j \\
    1
\end{bmatrix}.
\end{equation}

The rotation $(\alpha_{ij}, \beta_{ij})$ for the pixel $\mathbf{v}$ that corresponds to $\mathbf{p}$ becomes:
\begin{equation} \label{eq:theta_prime}
\alpha_{ij} = \arctan\left({p_x / \sqrt{p^2_x + p^2_y + p^2_x}}\right)
\end{equation}
\begin{equation} \label{eq:phi_prime}
\beta_{ij} = \arcsin\left({p_y / \sqrt{p^2_x + p^2_y + p^2_x}}\right).
\end{equation}

By using Equations \ref{eq:theta_prime} and \ref{eq:phi_prime}, we obtain the coordinate that corresponds to $\mathbf{v}$ in the equirectangular image:
\begin{equation} \label{eq:pano_coord_i}
u_i = (\alpha_{ij} + \pi) \times (W_{\text{equi}} / 2\pi)
\end{equation}
\begin{equation} \label{eq:pano_coord_j}
u_j = (\beta_{ij}  + \pi/2) \times (H_{\text{equi}} / \pi).
\end{equation}
Note that the pixel coordinates are obtained using bilinear interpolation.

The sizes of the equirectangular image for the SUN360 dataset and 360-Indoor are $(W_{\text{equi}}, H_{\text{equi}}) = (1024, 512)$ and $(W_{\text{equi}}, H_{\text{equi}}) = (1920, 960)$ respectively.
In our simulation, while the $f = 90^\circ$, the perspective image size is $(W_{\text{pers}}, H_{\text{pers}}) = (256, 256)$.
When $f = 60^\circ$, we used $(W_{\text{pers}}, H_{\text{pers}}) = (256, 192)$
Since this computation process needs to be fast enough for training and benchmarks, we implemented a GPU supported program for our task.

\section{Difficulties}

\subsection{Initial Conditions}
\label{apdx:initial_condition}

It is crucial in evaluating how the agent performs based on where the agent starts $\mathbf{R}_{\text{init}}$ in respect to the target view $\mathbf{R}_{\text{target}}$.
We denote the pitch and yaw rotations for initial and target view as $\mathbf{R}_{\text{init}}=(\theta_{\text{init}}, \psi_{\text{init}})$ and $\mathbf{R}_{\text{target}}=(\theta_{\text{target}}, \psi_{\text{target}})$ respectively.
We made three different initial conditions, `easy', `medium', `hard', which are described below:

\noindent\textbf{Easy.}
We categorize an episode as `easy' when the FoV overlap between each view satisfies the distance below:
\begin{equation} \label{eq:init_easy}
\| \mathbf{R}_{\text{target}} - \mathbf{R}_{\text{init}} \|_{2} \leq \frac{\sqrt{2} f}{2},
\end{equation}
where $f$ is the FoV in degrees.
We also note that $ N_{min} \times \delta \leq \| \mathbf{R}_{\text{target}} - \mathbf{R}_{\text{init}} \|_{1}$ is satisfied, where $N_{min}$ is the minimum number of steps for the episode and $\delta$ is the rotation increments in degrees.

\noindent\textbf{Medium.}
We categorize an episode as `medium' when the FoV overlap between each view satisfies the distance below:
\begin{equation} \label{eq:init_medium}
\frac{f}{2} < \| \mathbf{R}_{\text{target}} - \mathbf{R}_{\text{init}} \|_{1} \leq f.
\end{equation}
The views are located farther apart resulting in reduced overlaps.

\noindent\textbf{Hard.}
We categorize an episode as `hard' when the FoV overlap between each view satisfies the distance below:
\begin{equation} \label{eq:init_hard_1}
f < |\theta_{\text{target}} - \theta_{\text{init}}|,
\end{equation}
and
\begin{equation} \label{eq:init_hard_2}
f < | \psi_{\text{target}} - \psi_{\text{init}} |,
\end{equation}
where the two inequalities are satisfied jointly.
This ensures that the initial conditions have no overlaps.

The inequalities for determining difficulties are useful in randomly sampling views that satisfies the difficulties during training.
With these conditions, we created a difficulty scheduler for gradually introducing difficult episodes which ensures stability.

\subsection{Corruptions}
\label{apdx:corruption}

\begin{figure}[t]
\centering
\includegraphics[width=0.4\linewidth]{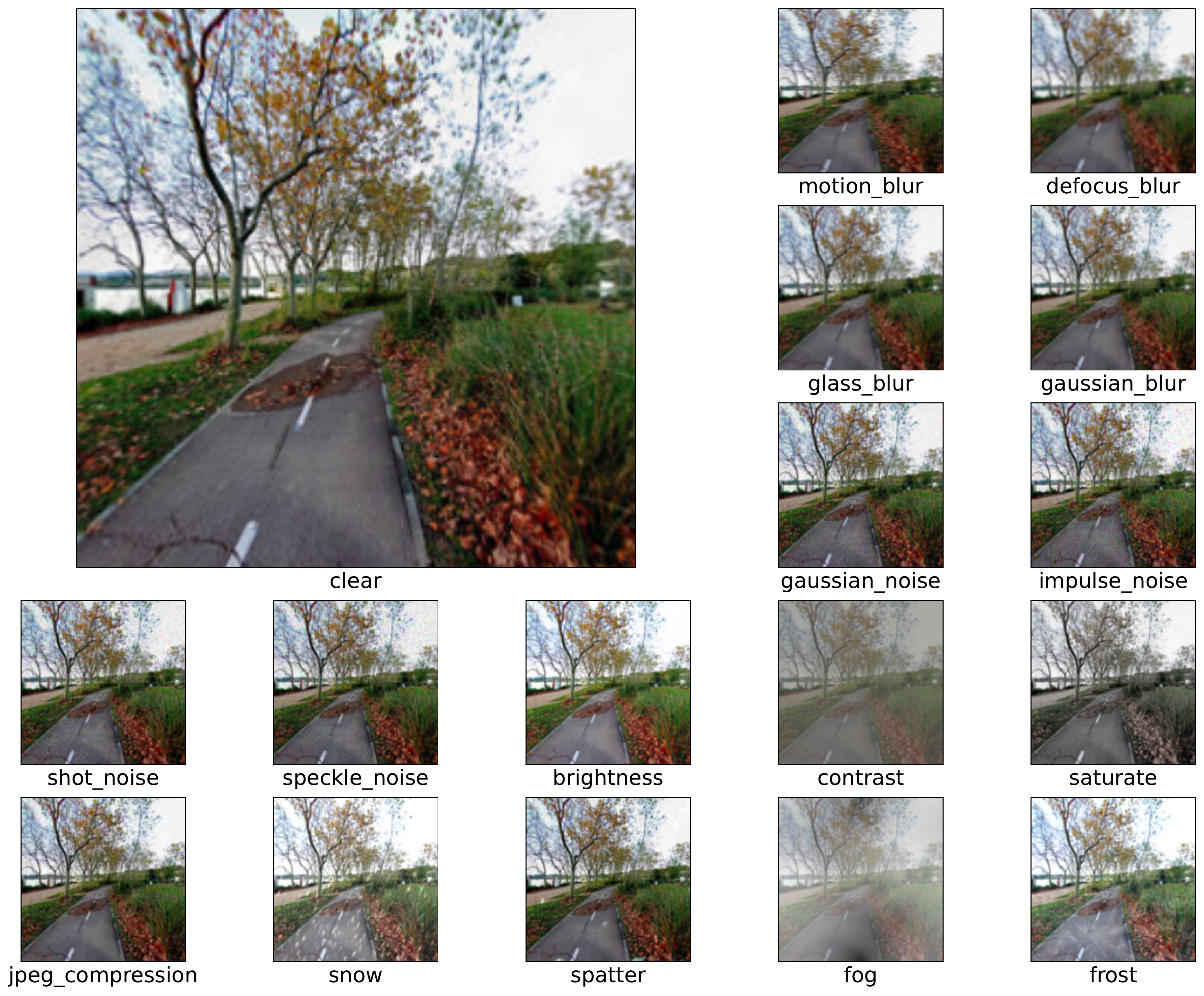}
\caption{Samples for low corruption (severity level of $1$).}
\label{fig:corruption_1}
\end{figure}

\begin{figure}[t]
\centering
\includegraphics[width=0.4\linewidth]{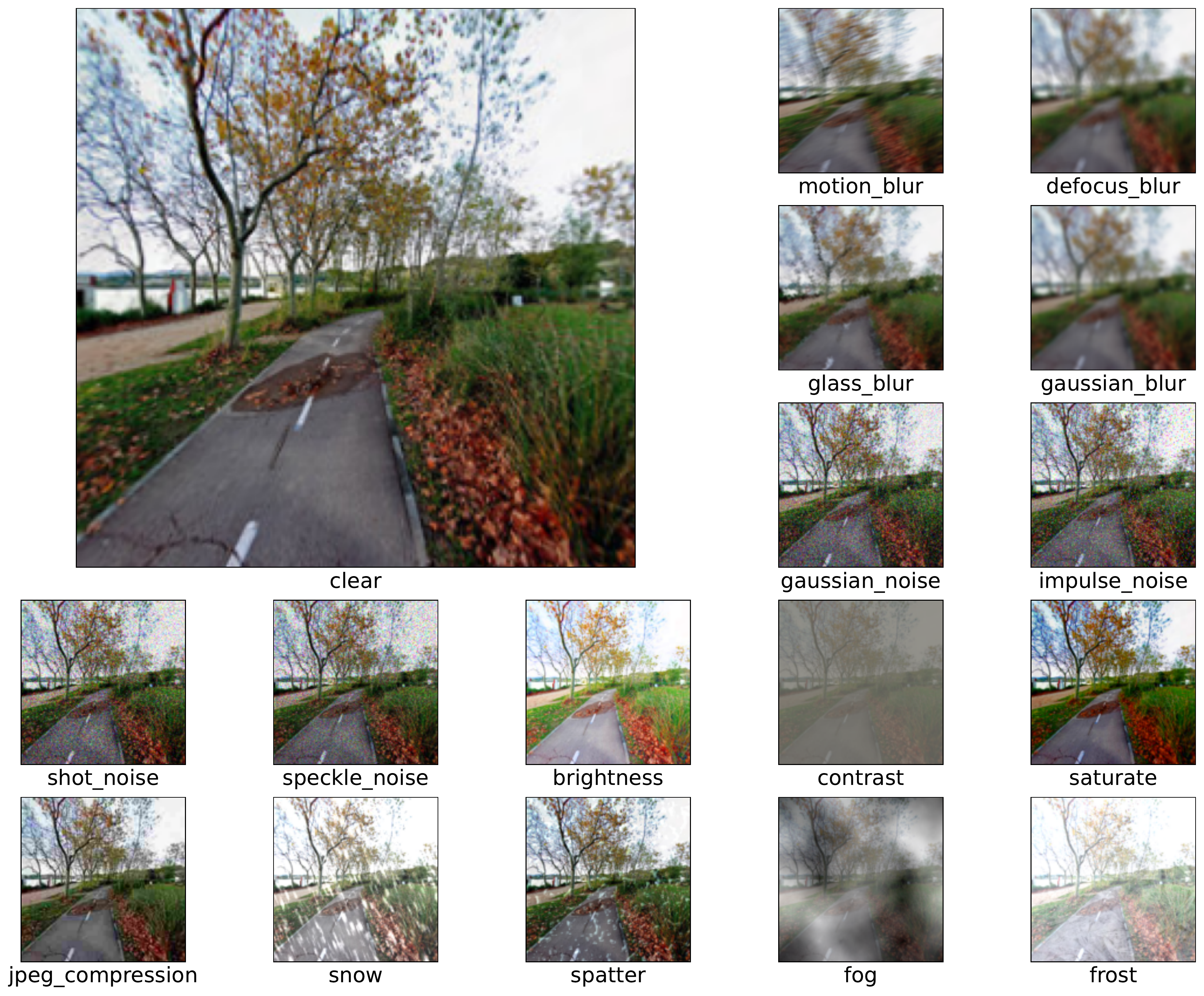}
\caption{Samples for high corruption (severity level of $3$).}
\label{fig:corruption_3}
\end{figure}

We visualize samples of low and high corruption on \cref{fig:corruption_1} and \cref{fig:corruption_3} respectively.
The corruptions are separated into four categories \cite{michaelis2019benchmarking} which are blur, noise, digital, and weather.
The blur corruptions consist of motion, defocus, glass, and Gaussian blur.
The noise corruptions consist of Gaussian, impulse, shot, and speckle noise.
The digital corruptions consist of brightness, contrast, saturation and JPEG compression.
The weather corruptions consist of snow, spatter, fog and frost.

\section{Details on the Policy Learning Approach}

\subsection{Synchronous Advantage Actor-Critic}
\label{apdx:a2c}

Generally, reinforcement learning methods can be grouped into two categories: value-based and policy-based methods.
Policy-based methods can directly optimize policy parameters $\theta$ to maximize the expected return \cite{SuttonBarto, policymethod}.
These methods update $\theta$ in the direction of $\nabla_\theta \log\pi_\theta(a_t | s_t)R_t$, where $\pi_\theta$ is a policy parameterized by $\theta$ and $R_t$ is the sum of expected future rewards.
Usually, the gradient computed from a few samples suffers from high variance.
REINFORCE family of algorithms \cite{REINFORCE}
reduce the variance by subtracting a baseline function, $b_t(s_t)$ form the expected return ($\nabla_\theta \log\pi_\theta(a_t | s_t)(R_t - b_t(s_t)$).
When the estimated advantage function $A(a_t, s_t) = Q(a_t, s_t) - V(s_t)$ is used for the critic and the actor is updated according to the oracle given by critic, 
the algorithm is called Advantage Actor-Critic. 
The Asynchronous Advantage Actor-Critic (A3C) algorithm \cite{mnih2016asynchronous} uses deep neural network to parameterize the policy as well as the value function, and replaces $R_t - b_t(s_t)$ in the policy gradient with the estimated advantage function. 
A3C runs multiple agents in parallel to collect the samples and asynchronously update the policy network parameters.

In our work, we used a synchronous version of A3C commonly known as A2C for training policies.
Compared to A3C which updates the policy network asynchronously, the advantage of the synchronous approach is that it allows for better utilization of GPUs for training since it can batch the roll-outs of trajectories \cite{wang2016learning}.

\subsection{Hyper-Parameters and Training Details}
\label{apdx:hyper_params}
For training the network, we used A2C to synchronously update the network parameters and Proximal Policy Optimization (PPO) \cite{PPO} to make the update more stable, as used in \cite{pytorchrl}. 
All models used in the experiments are trained using Adam with an initial learning rate of $0.0025$.
After every $128$ steps, epoch size for the PPO update is set to $4$ which uses a mini-batch size of $1$.
The value loss coefficient is $0.5$ and entropy coefficient is $0.01$.
We enable GAE and use linear learning rate and clip decays.
The discount factor is set to $\gamma = 0.99$ and the rest of the hyper-parameters used for training is the same as \cite{savva2019habitat}.

\subsection{Training Process}
\label{apdx:training}

\begin{figure}[t]
\centering
\includegraphics[width=0.4\linewidth]{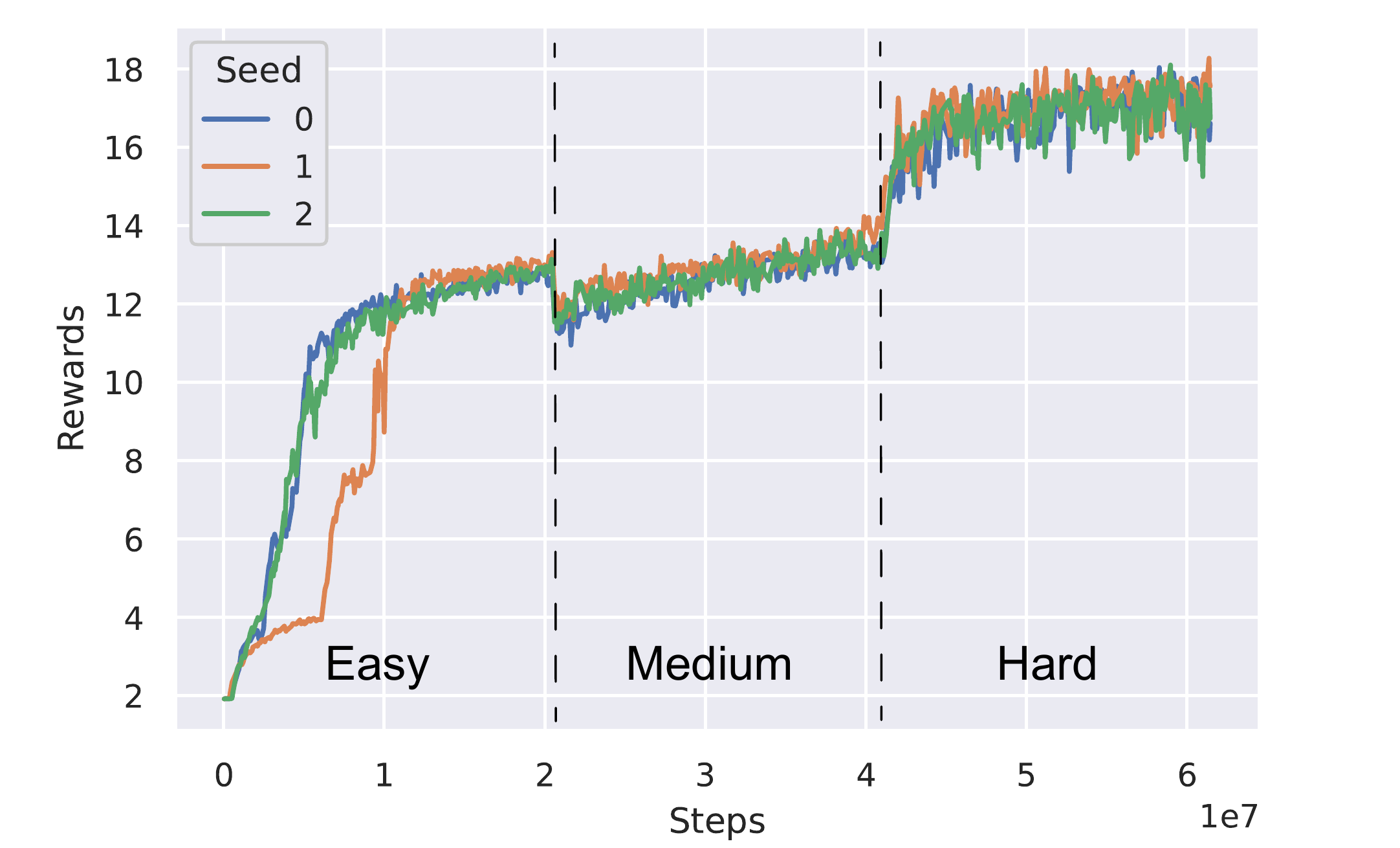}
\caption{Training curves of the \texttt{full} PPO agent.}
\label{fig:train_progress}
\end{figure}

Training the same agent with different seeds results in slightly different training curves as shown in \cref{fig:train_progress}.
We show the training progress of seeds $0$, $1$, and $2$.
The dotted lines represent the situations when the difficulty was raised using the difficulty scheduler, which is the same for all of the seeds.
We can see that for the `easy' difficulty, the agent quickly learns to optimize the reward function and settle around a reward of $12$.
As soon as the agents experience the `medium' difficulty, the rewards first reduce and then gradually increase to around $14$ as the agents experience more episodes.
Finally, when the agents experience `hard' difficulty, the rewards jumps to around $17$.
We believe that the significant jumps are caused by the agent learning to `look around' during the episodes during `medium' difficulty.
For episodes that are the `medium' episodes, the agent, by chance, could look in the optimal direction, but in the worse case, the agent might look the opposite direction, which results in lower rewards.
The big jump in reward could be explained by the agent who has learned to explore efficiently during the `medium' difficulty, and was able to optimize for `hard' episodes quickly.
Overall, despite the difference in the curves, the agents had arrived near the same rewards and thus we used seed $0$ for all of the training for the agents in our benchmarks.
We would like to conduct an extensive investigation for differing seeds in the future.

\section{Details on the Rule-based Approach}
\label{apdx:rulebased_details}

\begin{table}[t]
\tiny

\setlength\tabcolsep{3pt}
\centering
\caption{Validation Results of SUN360 Indoor.}
\begin{tabular}{ c c | c c c c c c c c c c c}
\hline\hline
Diff. & Agent & $10$ & $20$ & $30$ & $40$ & $50$ & $60$ & $70$ & $80$ & $90$ & $100$ & inf \T \\
\clineB{0-12}{1}
\multirow{2}{*}{Easy} 
  &\texttt{orb} & 19.4 & 8.11 & 1.02 & 0.44 & 0.36 & \textbf{0.24} & \textbf{0.24} & \textbf{0.24} & \textbf{0.24} & \textbf{0.24} & \textbf{0.24} \T \\
  &\texttt{sift}& 60.8 & 40.8 & 22.1 & \textbf{0.00} & \textbf{0.00} & \textbf{0.00} & 0.63 &\textbf{ 0.00} & \textbf{0.00} & 1.16 & 1.49 \T \\ \hline
\multirow{2}{*}{Medium} 
  &\texttt{orb} & 22.4 & 17.4 & 18.6 & 16.2 & 14.1 & 12.5 & 12.3 & \textbf{10.9} & \textbf{10.9} & \textbf{10.9} & \textbf{10.9} \T \\
  &\texttt{sift}& 78.7 & 54.2 & 21.9 & 8.83 & \textbf{2.50} & 3.32 & 8.84 & 9.93 & 9.18 & 7.51 & 12.6 \T \\ \hline
\multirow{2}{*}{Hard} 
  &\texttt{orb} & 32.0 & \textbf{21.1} & 26.8 & 40.0 & 40.8 & 33.4 & 37.1 & 38.5 & 38.5 & 38.5 & 38.5 \T \\
  &\texttt{sift}& 142.8 & 91.3 & 34.4 & 18.3 & \textbf{16.2} & 23.9 & 32.2 & 39.5 & 43.3 & 62.5 & 83.1 \T \\
\hline\hline
\end{tabular}
\label{table:validation_sun360_indoor}

\setlength\tabcolsep{3pt}
\centering
\caption{Validation Results of SUN360 Outdoor.}
\begin{tabular}{ c c | c c c c c c c c c c c}
\hline\hline
Diff. & Agent & $10$ & $20$ & $30$ & $40$ & $50$ & $60$ & $70$ & $80$ & $90$ & $100$ & inf \T \\
\clineB{0-12}{1}
\multirow{2}{*}{Easy} 
  &\texttt{orb} & 17.2 & 8.55 & 4.50 & 5.67 & 5.02 & 3.09 & \textbf{3.05} & \textbf{3.05} & \textbf{3.05} & \textbf{3.05} & \textbf{3.05} \T \\
  &\texttt{sift}& 51.9 & 35.0 & 16.1 & 4.43 & 1.64 & 1.20 & \textbf{0.00} & \textbf{0.00} & \textbf{0.00} & \textbf{0.00} & \textbf{0.00} \T \\ \hline
\multirow{2}{*}{Medium} 
  &\texttt{orb} & 27.8 & 16.9 & \textbf{6.82} & 7.35 & 13.2 & 16.1 & 16.1 & 16.1 & 16.1 & 16.1 & 16.1 \T \\
  &\texttt{sift}& 90.2 & 54.6 & 18.2 & 7.32 & 5.30 & 4.07 & 2.52 & \textbf{1.33} & \textbf{1.33} & \textbf{1.33} & 6.52 \T \\ \hline
\multirow{2}{*}{Hard} 
  &\texttt{orb} & 23.7 & 19.1 & 10.7 & \textbf{8.48} & 17.1 & 16.5 & 16.3 & 16.3 & 16.3 & 16.3 & 16.3 \T \\
  &\texttt{sift}& 141.1 & 86.5 & 33.6 & \textbf{3.19} & 3.34 & 8.22 & 10.9 & 11.0 & 14.5 & 27.7 & 72.8 \T \\
\hline\hline
\end{tabular}
\label{table:validation_sun360_outdoor}

\setlength\tabcolsep{3pt}
\centering
\caption{Validation Results of 360-Indoor.}
\begin{tabular}{ c c | c c c c c c c c c c c}
\hline\hline
Diff. & Agent & $10$ & $20$ & $30$ & $40$ & $50$ & $60$ & $70$ & $80$ & $90$ & $100$ & inf \T \\
\clineB{0-12}{1}
\multirow{2}{*}{Easy} 
  &\texttt{orb} & 17.6 & \textbf{8.55} & \textbf{8.55} & 12.3 & 9.57 & 11.4 & 11.4 & 11.4 & 11.4 & 11.4 & 11.4 \T \\
  &\texttt{sift}& 60.4 & 43.4 & 28.3 & 11.1 & 6.94 & 5.35 & 5.43 & \textbf{5.02} & 5.28 & 6.43 & 5.37 \T \\ \hline
\multirow{2}{*}{Medium} 
  &\texttt{orb} & 24.2 & 18.6 & \textbf{12.1} & 17.7 & 21.2 & 23.2 & 20.5 & 20.5 & 20.5 & 20.5 & 20.5 \T \\
  &\texttt{sift}& 93.6 & 70.7 & 35.0 & 15.4 & \textbf{9.47} & 13.2 & 15.3 & 16.6 & 20.0 & 20.0 & 24.3 \T \\ \hline
\multirow{2}{*}{Hard} 
  &\texttt{orb} & 30.1 & \textbf{16.0} & 24.0 & 35.4 & 39.1 & 41.1 & 39.0 & 41.4 & 40.9 & 40.9 & 40.9 \T \\
  &\texttt{sift}& 140.1 & 119.3 & 68.2 & 26.0 & \textbf{17.8} & 21.3 & 29.1 & 39.6 & 41.4 & 51.0 & 80.7 \T \\
\hline\hline
\end{tabular}
\label{table:validation_360indoor}
\end{table}

\subsection{Pseudocode}
\label{apdx:pseudo_code}

\begin{algorithm*}[ht]
\caption{Consensus Algorithm (Python-like Pseudocode)}
\label{alg:consensus}
\SetAlgoLined
$d_{\text{thresh}}$ \Comment{Threshold for maximum discriptor distance}
\Function{Consensus($M$, $K_{\text{current}}$, $K_{\text{target}}$, $a_{t-1}$)}{
  $A \assign \var{[]}$\Comment{Assign empty list}
  \For{$\var{m}$ $\text{\bf{in}}$ $M$}{
    \If{$\var{m.distance} > d_{\text{thresh}}$}{
      $d_x, d_y \assign \FuncCall{CaculcateDisplacement}{$\var{m}$, $K_{\text{current}}$, $K_{\text{target}}$}$\;
      \Comment{Calcuate the pixel-wise displacement based on the $o^{\text{current}}_t$.}
      \Comment{Note that $d_x$ and $d_y$ are scalar values that could be negative.}
      \eIf{$d_x ~ 0$ $\text{\bf{and}}$ $d_y ~ 0$}{
        $\var{a} \assign \texttt{"stop"}$\;
      }{
        \eIf{$\FuncCall{abs}{$d_x$} > \FuncCall{abs}{$d_y$}$}{
          \eIf{$d_x > 0$}{
            $\var{a} \assign \texttt{"right"}$\;
          }{
            $\var{a} \assign \texttt{"left"}$\;
          }
        }{
          \eIf{$d_y > 0$}{
            $\var{a} \assign \texttt{"up"}$\;
          }{
            $\var{a} \assign \texttt{"down"}$\;
          }
        }
      }
      $A += \var{[a]}$\Comment{Append to list}
    }
  }
  \If{$\FuncCall{len}{$A$} == 0$}{
    \Return{$a_{t-1}$}
  }
  $a_t \assign \FuncCall{mode}{$A$}$\;
  \Return{$a_t$}
}
\end{algorithm*}

\begin{algorithm*}[ht]
\caption{Single Loop of the Rule-based algorithm (Python-like Pseudo-code)}
\label{alg:action_estimation}
\SetAlgoLined
$N_{\text{kps}}$ \Comment{Threshold for number of keypoints}
$N_{\text{matches}}$ \Comment{Threshold for number of matches}
\Function{EstimateAction($o^{\text{target}}_t$, $o^{\text{current}}_t$, $a_{t-1}$)}{
  \Comment{Preprocess Input Images.}
  $o^{\text{target}}_t \assign \FuncCall{COLOR2Gray}{$o^{\text{target}}_t$}$\;
  $o^{\text{current}}_t \assign \FuncCall{COLOR2Gray}{$o^{\text{current}}_t$}$\;
  \Comment{Detect Keypoints and Compute Descriptors where $K$ and $D$ are lists.}
  $K_{\text{target}}, D_{\text{target}} \assign \FuncCall{detectAndCompute}{$o^{\text{target}}_t$}$\;
  $K_{\text{current}}, D_{\text{current}} \assign \FuncCall{detectAndCompute}{$o^{\text{current}}_t$}$\;
  \If{$\FuncCall{len}{$K_{\text{target}}$} < N_{\text{kps}}$ $\text{\bf{or}}$ $\FuncCall{len}{$K_{\text{target}}$} < N_{\text{kps}}$}{
    \Return{$a_{t-1}$}
  }
  \Comment{Match Features.}
  $M_{raw} \assign \FuncCall{KNNMatcher}{$K_{\text{current}}$, $K_{\text{target}}$}$\Comment{Match using KNN with $K=2$}
  $M \assign \var{[]}$\Comment{Assign empty list}
  \For{$i, \var{m}, \var{n}$ $\text{\bf{in}}$ $\FuncCall{enumerate}{$M_{raw}$}$}{
    \If{$\var{m.distance} < 0.7 \times \var{n.distance}$}{
      $M += \var{[m]}$\Comment{Append to list}
    }
  }
  \If{$\FuncCall{len}{$M$} < N_{\text{matches}}$}{
    \Return{$a_{t-1}$}
  }
  \Comment{Consensus Algorithm.}
  $a_t \assign \FuncCall{Consensus}{$M$, $K_{\text{current}}$, $K_{\text{target}}$, $a_{t-1}$}$\Comment{See \ref{alg:consensus}}
  \Comment{Detect if the agent is performing oscillation-like trajectory.}
  \If{$\FuncCall{isRepeated}{$a_t$}$}{
    \Return{$\texttt{"stop"}$}
  }
  \Return{$a_t$}
}
\end{algorithm*}

In \ref{alg:action_estimation}, we show a rough, Python-like pseudo-code for the rule-based agent.
As prerequisites, we implemented the algorithm using NumPy and OpenCV in Python, therefore most of the functions are APIs from OpenCV.
For our algorithm, we initialize the feature detector and KNN matching algorithm using the $\FuncCall{detectAndCompute}{}$ and $\FuncCall{KNNMatcher}{}$ functions respectively.
The minimum number of keypoints is $N_{\text{kps}} = 500$ and the minimal number of matches is $N_{\text{matches}} = 10$.
These parameters are not descriptor specific.
In our work, we used two different types of descriptors, ORB and SIFT \cite{lowe2004distinctive,rublee2011orb}.
From preliminary tests, we found that $d_{\text{thresh}}$ affects the localization accuracy depending on the descriptors and difficulties.
In \ref{apdx:param_search}, we describe the method to obtain the best values of $d_{\text{thresh}}$ using the validation set.

We offer a brief explanation of how the algorithm works.
At step $t$, $\FuncCall{EstimateAction}{}$ takes in $o_t = (o^{\text{target}}_t, o^{\text{current}}_t)$ and the previous action $a_{t-1}$.
Note that $o^{\text{current}}_t$ is the current perspective image in the environment at step $t$.
First, the observations were converted into grayscale images.
Then, we detected keypoints and their descriptors.
Note that if we have smaller number of keypoints than $N_{\text{kps}}$, the previous action is returned.
Next, we matched the two lists of keypoints to obtain rough matches.
We refined the matches through ratio tests following \cite{lowe2004distinctive}.
If the resulting list of matches $M$ is smaller then $N_{\text{matches}}$, the previous action is returned.
We used a consensus algorithm to estimate the best action, described in \ref{alg:consensus}.
For this algorithm, we calculated the pixel-wise displacement for each match and estimated the direction to move.
The mode of estimated actions was returned as $a_t$.
Finally, we kept track of each action $a_{i}$ for $i \in [0, t]$ and checked for unexpected behaviors.
For example, we found that agents will oscillate at a certain point when they reach some local minima and would not call \textit{"STOP"}.
We excluded such behaviors in this post-processing and force the agent to stop early.

During observations with low keypoints or matches, the agent is programmed to look in either \texttt{left} or \texttt{right} and would continue looking in the same direction until a feature is found.

\subsection{Parameter Search}
\label{apdx:param_search}

We have performed parameter search for $d_{\text{thresh}}$ using the validation set of the datasets.
We evaluated each agent with different $d_{\text{thresh}}$ values for each difficulty.
The threshold values we investigated were $10$, $20$, $30$, $40$, $50$, $60$, $70$, $80$, $90$, $100$, and `infinity' (which represents no threshold).
In the evaluation, we find the value of $d_{\text{thresh}}$ that minimizes localization error $\varepsilon$.
\cref{table:validation_sun360_indoor,table:validation_sun360_outdoor} shows the results of the agents evaluated in SUN360 Indoor and Outdoor datasets respectively.
\cref{table:validation_360indoor} shows the results of the agents evaluated in 360-Indoor.

As shown in \cref{table:validation_sun360_indoor}, the values in bold represent the lowest localization error $\varepsilon$.
The threshold values were used in the benchmarks for SUN360 Indoor datasets in \cref{sec:sun360_experiment}.
Note that that values differ in difficulties.
And in the benchmarks, we also used a threshold value that is specific to their difficulty.
The values shown in \cref{table:validation_sun360_outdoor} were used to benchmark SUN360 Outdoor dataset.
Note that these deduced threshold values are also used for the corruption benchmarks in \cref{sec:corruptions}.

The validation results for 360-Indoor dataset is shown in \cref{table:validation_360indoor}.
Similarly with other validation results, the thresholds that minimizes localization error $\varepsilon$ are used for the benchmarks in \cref{sec:360indoor_experiment}.
Besides `hard', it is interesting that the threshold values differ from SUN360 Indoor even though the domain is fairly close.

Note that there were sometimes multiple values for $d_{\text{thresh}}$ that minimize $\varepsilon$, but for the benchmarks and evaluations, we have used one of the values.
The agents that have the exact $\varepsilon$ also had the exact values for the other metrics, down to the same trajectories.

\section{Implementation Details}
\label{apdx:implementation details}

All of the training were run on machines with Intel(R) Xeon(R) Gold 5220R CPU @ 2.20GHz with NVidia RTX 5000 GPU.
All of the benchmarks were run on machines with AMD Ryzen 9 3900X 12-Core Processor with NVidia RTX 3090 GPU.
The code was implemented using Python with the models and training based on the PyTorch framework.
\texttt{orb} and \texttt{sift} has OpenCV functions.
We encourage the readers to see the supplemented code for more details.

Indoor and Outdoor splits for SUN360 dataset will be publicly available for download.
360-Indoor splits will also be publicly available.
The source equirectangular images cannot be redistributed, therefore not included in the code.
To reproduce the results, one must agree to the guidelines and obtain permission to download the source images from the owners of the SUN360 or the 360-Indoor datasets \cite{xiao2012recognizing,chou2020360}.



\end{appendices}

\bibliographystyle{unsrtnat}
\bibliography{references}  

\end{document}